\documentclass[a4paper, 10pt, conference]{ieeeconf}      %

\IEEEoverridecommandlockouts                              %

\usepackage[backend=bibtex,bibstyle=ieee,citestyle=numeric-comp,maxbibnames=3,maxcitenames=3,doi=false,url=false,eprint=false]{biblatex}
\usepackage{graphics} %
\usepackage[pass]{geometry} %
\usepackage{graphicx}
\usepackage{booktabs} %

\usepackage[svgnames]{xcolor}
\usepackage{amsmath}
\usepackage{svg}
\usepackage{subcaption}
\usepackage{amssymb}
\usepackage{float}
\usepackage{multirow}
\usepackage{siunitx}

\newcommand{\cn}[1][]{\textsuperscript{\textcolor{red}{\ifthenelse{\equal{#1}{}}{[citation needed]}{[citation needed: {#1}]}}}}
\usepackage{hyperref}

\usepackage[nameinlink]{cleveref}
\Crefname{equation}{Eq.}{Eq.}
\Crefname{figure}{Fig.}{Fig.}
\Crefname{tabular}{Tab.}{Tab.}
\Crefname{section}{Sec.}{Sec.}

\title{\LARGE \bf
CoCar NextGen: a Multi-Purpose Platform for\\Connected Autonomous Driving Research
}

\author{Marc Heinrich$^{1}$, 
Maximilian Zipfl$^{1}$, 
Marc Uecker$^{1}$,
Sven Ochs$^{1}$,
Martin Gontscharow$^{1}$, \\
Tobias Fleck$^{1}$,
Jens Doll$^{1}$,
Philip Sch\"orner$^{1}$,
Christian Hubschneider$^{1}$,
Marc Ren\'{e} Zofka$^{1}$,\\
Alexander Viehl$^{1}$
and J. Marius Z\"ollner$^{1,2}$%
\thanks{$^{1}$ Department of Technical Cognitive Systems, FZI Research Center for Information Technology, Germany.
	{\tt\small \{surname\}@fzi.de}}%
\thanks{$^{2}$ Karlsruhe Institute of Technology (KIT), Germany.}
}

\begin{document}

\maketitle
\thispagestyle{empty}
\pagestyle{empty}

\begin{abstract}

Real world testing is of vital importance to the success of automated driving. While many players in the business design purpose build testing vehicles, we designed and build a modular platform that offers high flexibility for any kind of scenario. CoCar NextGen is equipped with next generation hardware that addresses all future use cases. Its extensive, redundant sensor setup allows to develop cross-domain data driven approaches that manage the transfer to other sensor setups. Together with the possibility of being deployed on public roads, this creates a unique research platform that supports the road to automated driving on SAE Level 5.

\end{abstract}

\section{INTRODUCTION}

Autonomous driving test vehicles (AVs) are at the vanguard of innovation in autonomous mobility. These vehicles are equipped with cutting-edge sensors; Cameras, LiDAR, radar, and sophisticated software enable them to perceive and interpret their surroundings with unmatched precision. Beyond their hardware and software components, the significance of test vehicles lies in their role as real-world laboratories. They provide an essential bridge between theoretical development and practical application, offering a controlled environment in which autonomous driving systems can be rigorously evaluated, refined, and validated. The system under test needs a driving platform to be evaluated under real-world conditions. The system encompasses a range of elements, including individual components like sensors and innovative machine learning approaches, such as those related to perception or planning components, up to entire automated driving stacks.

\begin{figure}[t]
    \includegraphics[width=\columnwidth]{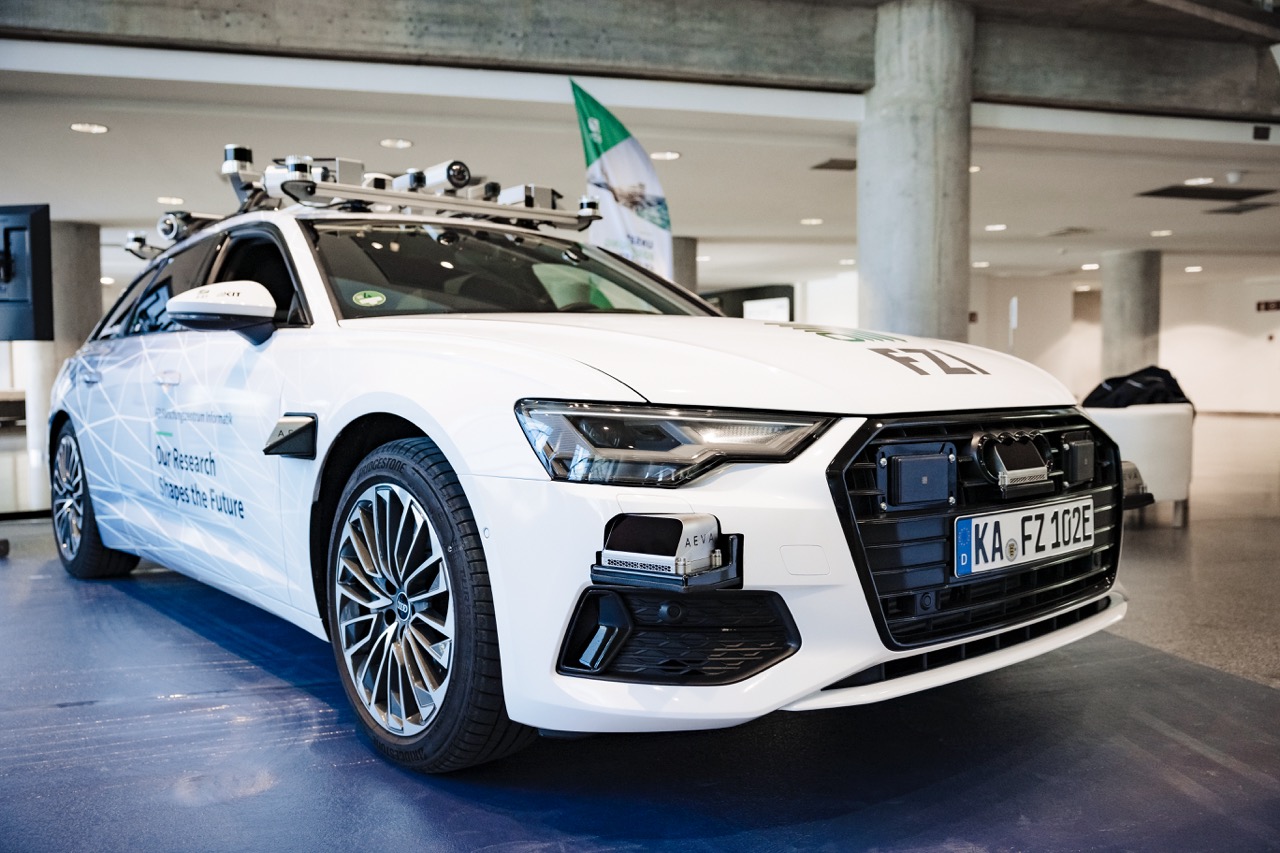}
    \label{fig:cc-ng-front}
    \caption{CoCar NextGen was presented first on the IEEE ITSC 2023 in Bilbao}
\end{figure}

While large manufactuers are able to deploy multiple fleets of testing vehicles built for different applications, an independent research institution like the FZI Forschungszentrum Informatik needs to operate a single vehicle to address all research scenarios. Therefore, we designed and built a vehicle with maximum flexibility in mind. The focus was on modular and standardised interfaces. Use cases we have considered include, but are not limited to:
\begin{itemize}
    \item Recording of real world data
    \item Closed loop testing of automated driving software
    \item Communication with connected intelligent infrastructure
    \item Studies on acceptance and user experience
\end{itemize}
The variety of use cases demand an extensive hardware setup. Therefore, this vehicle must not be understood as a minimal setup to conduct automated driving, rather than a platform for future research. Moreover, the extensive setup provides unique opportunity for cross-domain research of multi-modal sensor setups.

The following sections provide an overview of the current research vehicles used in industrial and research contexts.  Subsequently, we introduce our latest experimental platform, CoCar Nextgen, starting with the design process following with a detailed description of the components.
Finally, we provide vehicle data alongside sample sensor data.

\section{RELATED WORK}
The industry of highly automated vehicles has undergone rapid changes in recent years. New technologies are not only being researched by universities, traditional vehicle manufacturers and component suppliers. The increased public interest in this field has also sprouted a healthy ecosystem of start-ups which are also contributing to different research areas.
Large technology companies including Alphabet, Apple, and Amazon, have also invested into subsidiaries in this sector with the aim of developing self-driving vehicles.

In this domain of research, a spectrum of different approaches have emerged for building vehicles and sensor setups.
Traditional vehicle manufacturers, suppliers and profit-oriented startups often focus on leaner vehicle setups. Their prototypes aim towards resembling production vehicles with concern for transition into series production at fleet scale, and are often bound to immediate business cases.
An extreme example of this approach is seen in Tesla's vehicle setup, which have completely eliminated LiDAR and RADAR sensors in favor of a cheaper multi-camera surround setup in order to cut costs \cite{tesla, tesla_autopilot, tesla_owners_manual}.
However, even vehicle manufacturers and suppliers maintain a smaller set of internal one-off prototypes for pre-production research involving less mainstream technologies and concepts. Unfortunately, the full details are rarely published.

Much more is known about the well-funded startups, such as Zoox, Waymo, or Cruise.
These companies often proudly present selected parts of their vehicles and technologies, including their vehicle setups \cite{waymo,gm_self_driving_report_18}.
These vehicles feature a multitude of different sensors in configurations focused on \ang{360} coverage of the surroundings with different sensor types, typically including LiDAR, RADAR and cameras~\cite{chitlangia2021improving}. These vehicles serve a combined purpose of being both a prototype and testing ground, but also a means of data collection. Based on their public-facing use in robotaxi services, these vehicles are also mostly geared towards economically scaling into fleet deployment.

At the other end of the spectrum, research vehicles built by universities and publicly-funded non-profit institutions often serve more basic research, which is under less pressure to immediately be useful to a business case. As a result, these vehicles are often built with different use cases in mind. For instance, AnnieWAY, the vehicle used to record the commonly-used KITTI dataset \cite{sommer_team_2008,geiger_vision_2013} was built with a clear focus on data collection for perception and localization tasks. Its setup includes a stereo camera setup, as well as a single \ang{360} LiDAR sensor.
Similarly, Bertha, a joint research vehicle by Karlsruhe Institute of Technology and Daimler\cite{dang_autonomes_2015} was initially built for an autonomous driving demonstration, with the goal of driving a pre-defined route of 80 kilometers without driver intervention. Based on their use case, Bertha was equipped with a \ang{360} LiDAR sensor, as well as multiple forward- and a single rear-facing camera, which were used for environment perception, traffic sign detection and localization respectively.

Both AnnieWAY and Bertha were built specifically to solve the challenge they competed in. Therefore their hardware setup is tailored to the task, which does not support a broad research across multiple scenarios and domains.
In contrast to the previously mentioned, mostly application-specific, vehicle setups, EDGAR \cite{karle_edgar_2023}, a vehicle of the Technical University of Munich, was developed as a general-purpose research platform for automated driving. Therefore, EDGAR is likely the closest vehicle in scope to the vehicle presented in this paper.
The sensor setup is comprised of multiple LiDAR and RADAR sensors, a GPS/IMU unit, as well as multiple cameras. Moreover, microphones are also installed.
Six cameras are installed around the roof of the vehicle, and two long-range cameras are aligned towards the front, providing stereo vision. Two depth cameras are also positioned at the front of the vehicle. The LiDAR setup comprises four sensors, including two \ang{360} LiDARs on either side of the vehicle roof, each tilted slightly outwards. Additionally, two solid-state LiDAR sensors are aligned to the front and rear of the vehicle respectively, intended to detect distant objects at a range of up to up to \SI{250}{\meter}.
The camera and LiDAR setup has been enhanced with radar sensors to improve perception in severe weather conditions. Six radars are installed to cover a 360° range, alternating between long and short range measurements to detect objects at great distances. A digital twin of EDGAR was created to facilitate simulation driven verification and validation. The EDGAR vehicle is also equipped with a drive-by-wire system to facilitate testing end-to-end driving functions in real-world conditions.

As comprehensive as the EDGAR vehicle is, there are still some use cases for which it is only suitable to a limited extent. While it includes a desktop computer system for software-based driving functions, sustaining sufficient computational capabilities for multiple highly computationally-intensive tasks such as machine-learning based driving components may be difficult for this vehicle.
Additionally, a \ang{360} surround coverage of the environment is only provided by combining the different sensor modalities of EDGAR. The vehicle setup also only provides a heterogeneous mix of sensor types for full coverage of the surroundings.
LiDAR sensors currently pose the most promising candidate for privacy-preserving machine learning based environment modeling methods. Continued research in this domain, especially for transitioning the field from individual datasets to real-world vehicle deployment is crucial.
However, the cost of annotating 3D sensor data for accurate 3D environment perception is immense.
To overcome this problem of limited training data availability, multiple methods for cross-sensor and cross-modality labeling of sensor data have been developed in the past \cite{uecker2021, Piewak_2018, rist}. For these methods, the EDGAR vehicle may not be suitable, as they require large overlaps in sensor field of views, with ideally minimal parallax between sensor modalities. Similar sensor overlap requirements are often seen in low-level sensor fusion tasks, where projecting LiDAR point clouds into camera images is often a necessary intermediate step.

A further innovation which has recently joined the market, but is not yet integrated into the EDGAR vehicle, are LiDAR sensors which directly measure velocity estimates through Doppler effect measurements. These sensors could be a promising avenue towards improved localization, perception and object tracking methods.

\section{DESIGN PROCESS}
\label{sec:design}
As a research institution, our work ranges across the entire range of sub-systems in autonomous driving, including localization, perception, prediction, planning, and control. Due to this reason, we have a wide variety of use cases to cover for this research vehicle.

The use cases include recordings of sensor data, open loop testing of single software components, and closed loop testing of entire software stacks. Additionally, cooperative driving with other intelligent traffic participants and infrastructure is conducted. Finally, the quality of the equipment should be sufficient to serve as a reference for other systems to compare against.
Our driving scenarios range from densely populated urban environments via country roads to highway driving at high velocities. Additionally, applications such as valet parking in garages are also important. These environments dictate a versatile sensor setup, that excels both in range and in dealing with blind spots and occlusions. Moreover, the vehicle needs to locate itself without GNSS signal coverage. We aim to operate these scenarios in all environment conditions, including rain, fog and night. Finally, a road approval is mandatory.
The software that will be deployed includes both high performance applications, as well as machine learning algorithms. Moreover, our software stack consists of modular nodes that can be exchanged easily. Therefore, a flexible and powerful computing platform is needed, that is also capable of processing the incoming sensor data. To be able to deploy arbitrary software, the computing platform should be general purpose rather than dedicated embedded hardware.
Our research process aims for agile development cycles. Often, improvements to the software are implemented in the vehicle. Thus, comfort, usability and accessibility are important. Key factor is an easy-to-use interface.
Our goal was to design a vehicle that would cover all present and near-future research needs as far as possible. Therefore, the vehicle should be constructed in a modular way, such that future hardware can be added or upgraded easily.

The choice of the base vehicle is of high importance to the overall concept.
Our main objective was to have as much flexibility as possible.. Space was a top concern due to the need of fitting a variety of components. Conversely, our main deployment is in urban areas, where garages are as low as \SI{1.8}{\meter}. Consequently, SUVs are too high, often having heights of more than \SI{1.7}{\meter}, not leaving enough headroom to mount components on the roof.
To fulfill the requirement of a general purpose compute platform, we needed to fit a \SI{80}{\centi\meter} server rack into the trunk (see \Cref{ssec:compute}). Additionally, the vehicle needs to provide sufficient rear seat space to develop with notebooks. As sedans only provide limited vertical boot space, the choice was made for a large estate vehicle. An additional advantage is that the cabin air can be used to cool the components in the back, most notably the server.
Finally, the drive train was of importance. Our motion profile consists of many short distance trips. In development operations, there are also long periods of stationary operation when the software is adapted. This kind of operation ruled out diesel engines, as they suffer especially from short trips. Although considered, no battery electric vehicle (BEV) was available on the market at the time of the purchase as a large estate. Moreover, it was questionable if BEVs could provide sufficient power to support our component power system. To limit the effect on the environment, the choice was made for a plugin hybrid electric vehicle (PHEV). Finally, availability resulted in the choice of an Audi A6 Avant 50 TFSI e quattro. 

\subsection{Sensor Setup Design}
\label{ssec:sensor-setup-design}
\begin{figure*}[t]%
\begin{subfigure}[t]{0.247\linewidth}%
\includegraphics[width=\textwidth]{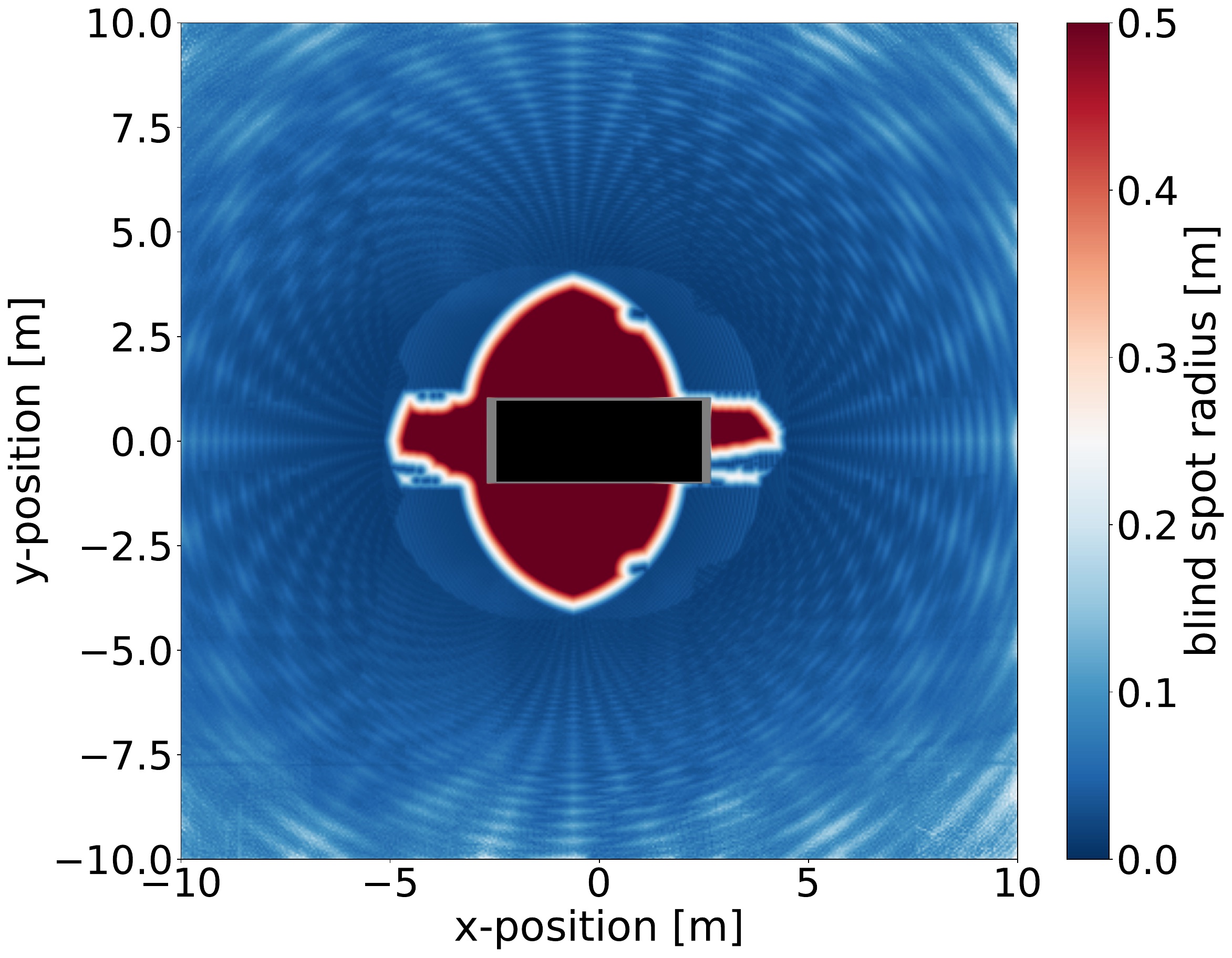}%
\caption{Mid-Range LiDAR}%
\end{subfigure}\hfill%
\begin{subfigure}[t]{0.247\textwidth}%
\includegraphics[width=\textwidth]{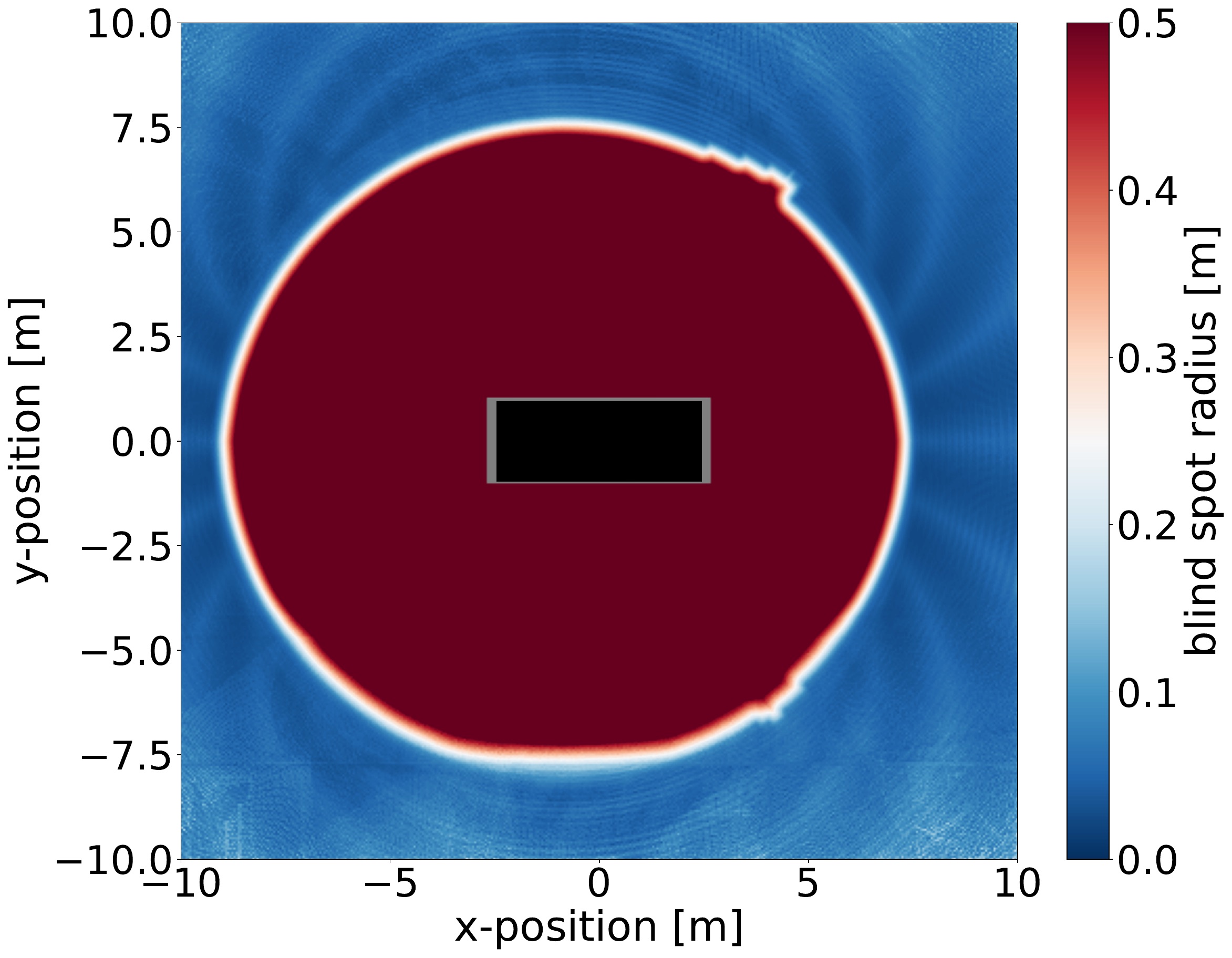}%
\caption{Long-Range  LiDAR}%
\end{subfigure}\hfill%
\begin{subfigure}[t]{0.247\textwidth}%
\includegraphics[width=\textwidth]{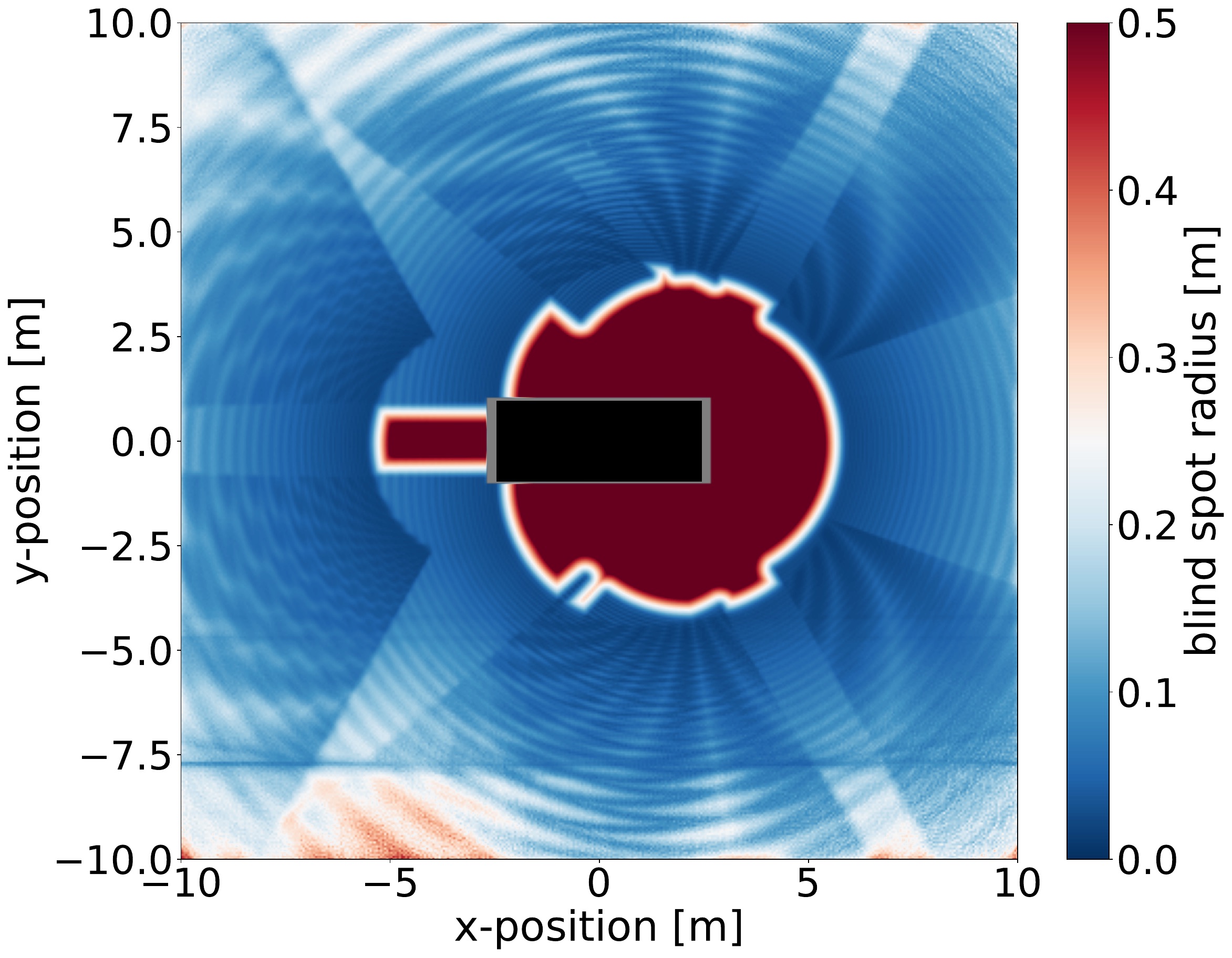}%
\caption{4D LiDAR}%
\end{subfigure}\hfill%
\begin{subfigure}[t]{0.247\textwidth}%
\includegraphics[width=\textwidth]{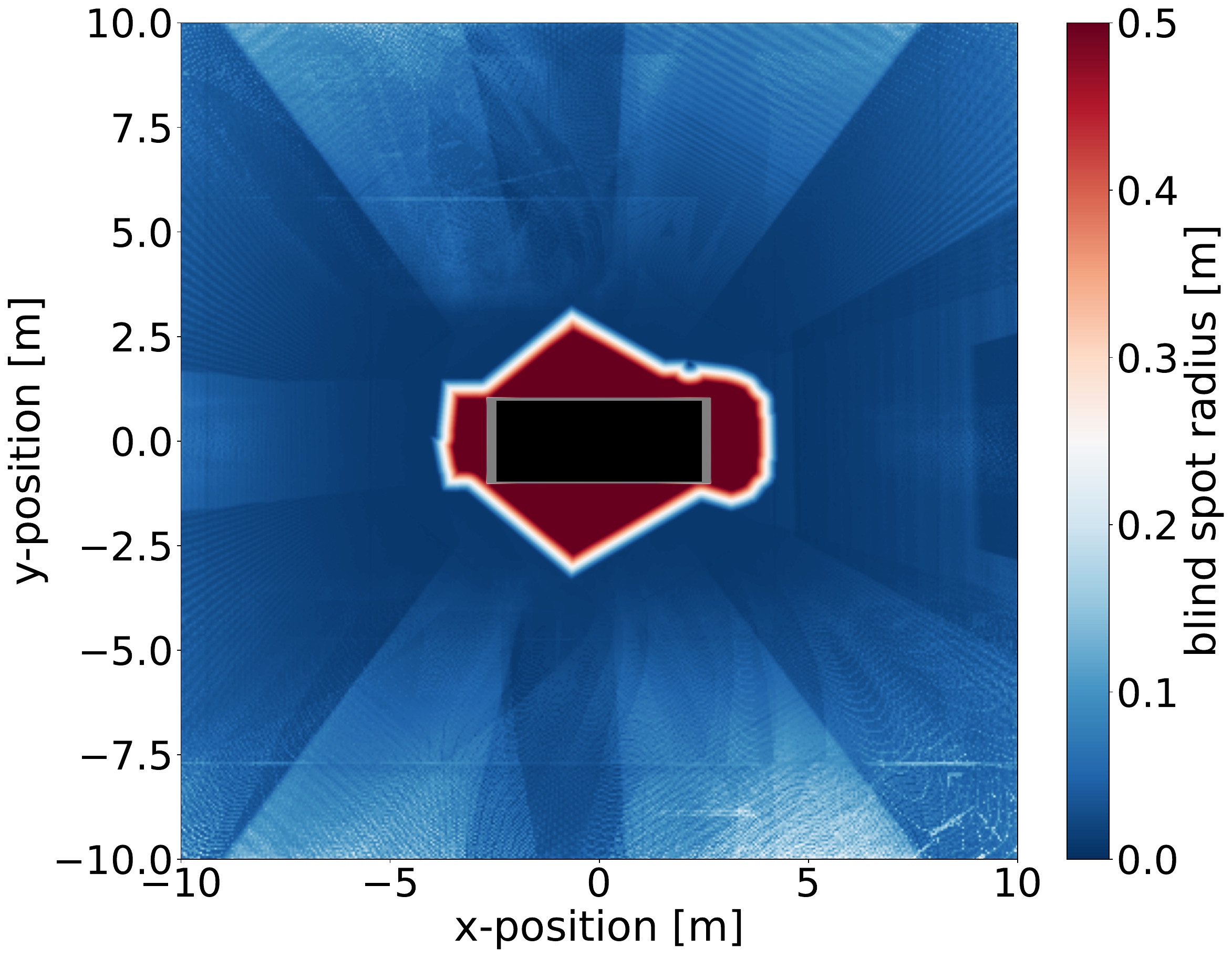}%
\caption{Cameras}%
\end{subfigure}%
\caption{Comparison of the ground level field-of-view of the different sensors at close range.}%
\label{fig:fov}
\end{figure*}

To design the sensor setup for our vehicle, we focused on providing a surround-view of the vehicle's environment as far as possible for each individual sensor modality.
To do this, we apply a recently-developed technique \cite{uecker2024} for estimating the blind spots and coverage of a sensor setup, based on simulation data from our digital twin.
As illustrated in \Cref{fig:fov}, each individual set of LiDAR and camera setups from our vehicle provides \ang{360} surround coverage of our vehicle's environment. During this development process, we aimed to minimize blind spots as far as possible. A different design decision was to co-locate multiple sensors on the roof to minimize occlusion from other traffic participants, as well as minimize parallax between sensors for improved LiDAR-to-camera projection for cross-sensor labeling or calibration tasks.

\section{HARDWARE SETUP}
\label{sec:sensors}

\begin{figure}[t]
\end{figure}

With our new vehicle, our goal was to create a platform that is able to cover all our research use cases. As opposed to a single use case vehicle, this general approach demands for an extensive hardware setup. To cover all use cases, we utilize different modalities of sensors. The modalities complement each other with their strengths. LiDAR offers direct and precise 3D representation of objects, while camera images provide semantic information and highest resolution. Finally, radar especially excels in adverse environmental conditions such as fog and heavy rain. To start with the exterior, our setup comprises the following components, as shown in \Cref{fig:cc-ng-rendering}: 

\begin{itemize}
    \item 6x - 4D-LiDAR-Scanner (cyan)
    \item 4x - Mid-Range \ang{360}-LiDAR (blue)
    \item 2x - Long-Range-\ang{360}-LiDAR (yellow)
    \item 9x - Full-HD-Camera (orange)
    \item 3x - 4D-Radar (green)
    \item 1x - High-Precision dual-antenna GNSS-System with IMU
    \item 1x – V2X Onboard Unit and 5G communication.
\end{itemize}
\Cref{fig:cc-ng-rendering} also indicates the positioning of the sensors. The majority of sensors have been integrated into a rooftop structure. This choice was made to optimize flexibility in accommodating sensor updates or incorporating additional sensors. Furthermore, this configuration simplifies maintenance. In the next section, we further motivate our decision-making regarding the sensor modalities and their respective positioning on the vehicle.

\subsection{LiDAR Sensors}
Our research group has a particular emphasis on LiDAR based perception. Different from an OEM building a fleet vehicle, our goal was less to create an economically feasible setup, but rather to achieve maximum flexibility. This is especially visible with our extensive LiDAR setup. In its entirety, the setup comprises 12 sensors that are integrated both within the vehicle body as well as onto the roof rack. The setup can be divided into three groups.

\begin{figure*}[t]
    \begin{subfigure}[b]{0.3\textwidth}
        \includegraphics[width=\textwidth]{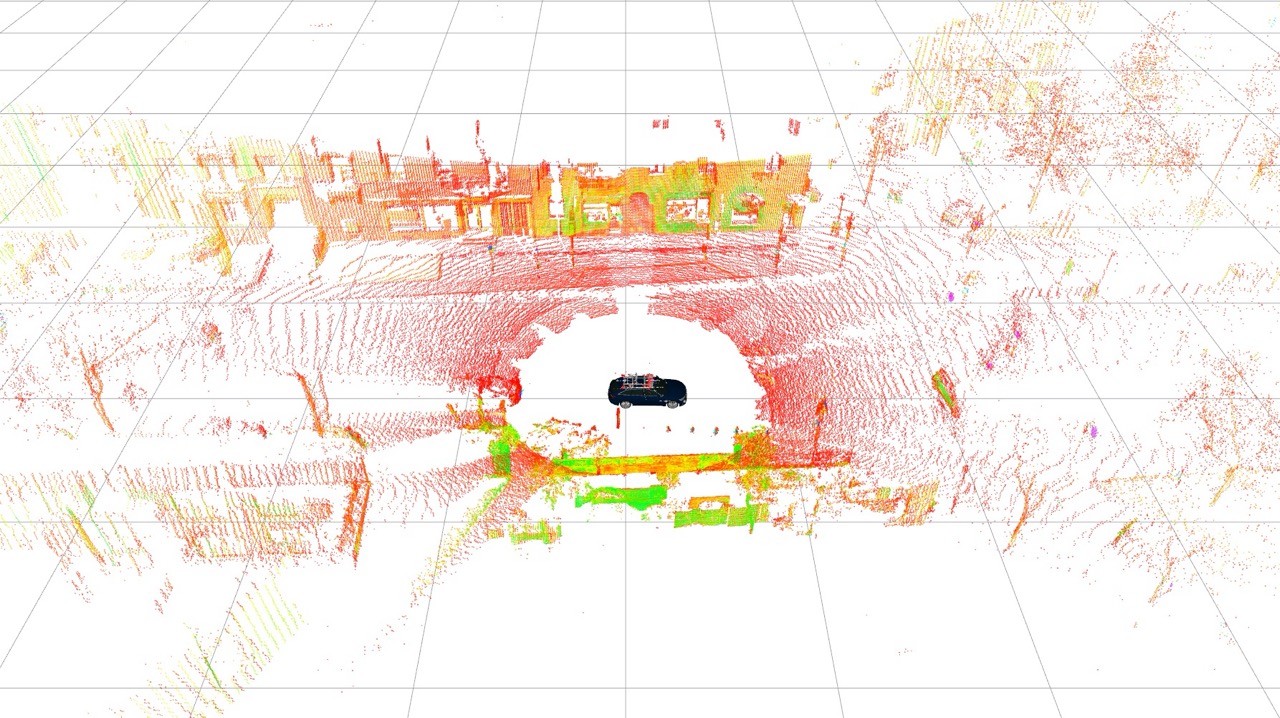}
        \caption{Long-Range LiDAR}
        \label{fig:os2}
    \end{subfigure}%
    \hspace{0.02\textwidth}
    \begin{subfigure}[b]{0.3\textwidth}
        \includegraphics[width=\textwidth]{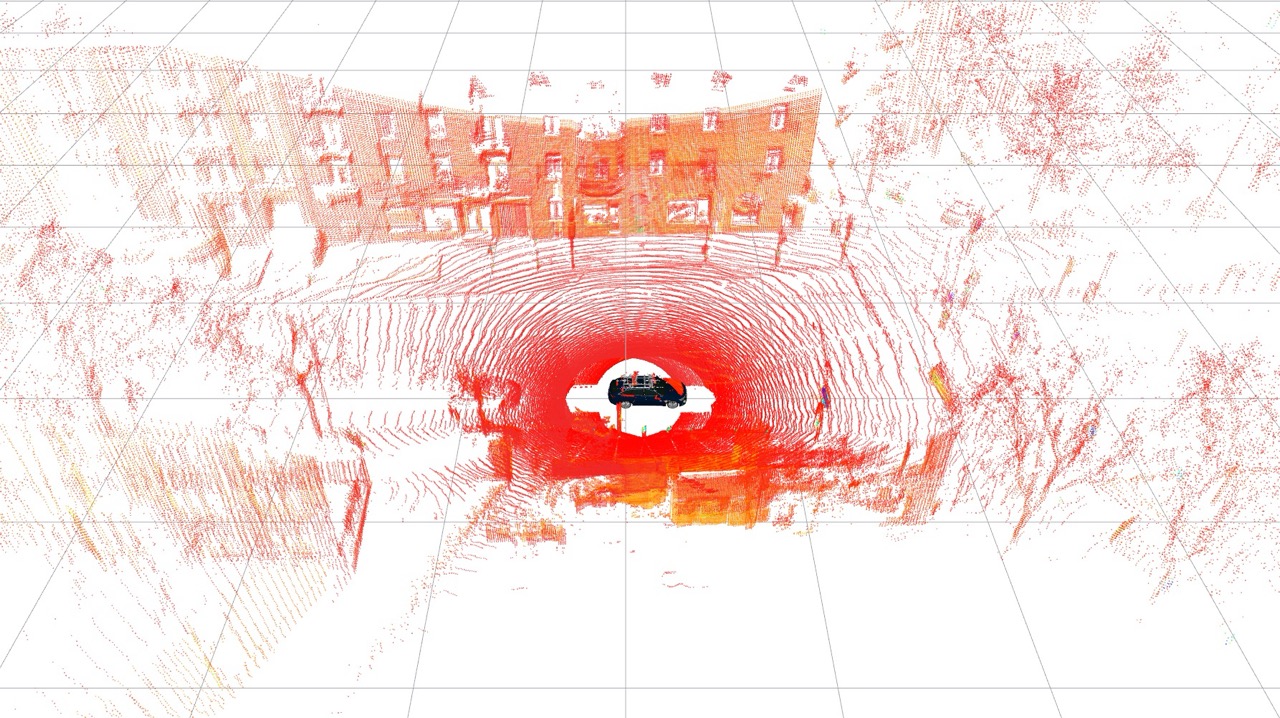}
        \caption{Mid-Range LiDAR}
        \label{fig:os1}
    \end{subfigure}%
    \hspace{0.02\textwidth}
    \begin{subfigure}[b]{0.3\textwidth}
        \includegraphics[width=\textwidth]{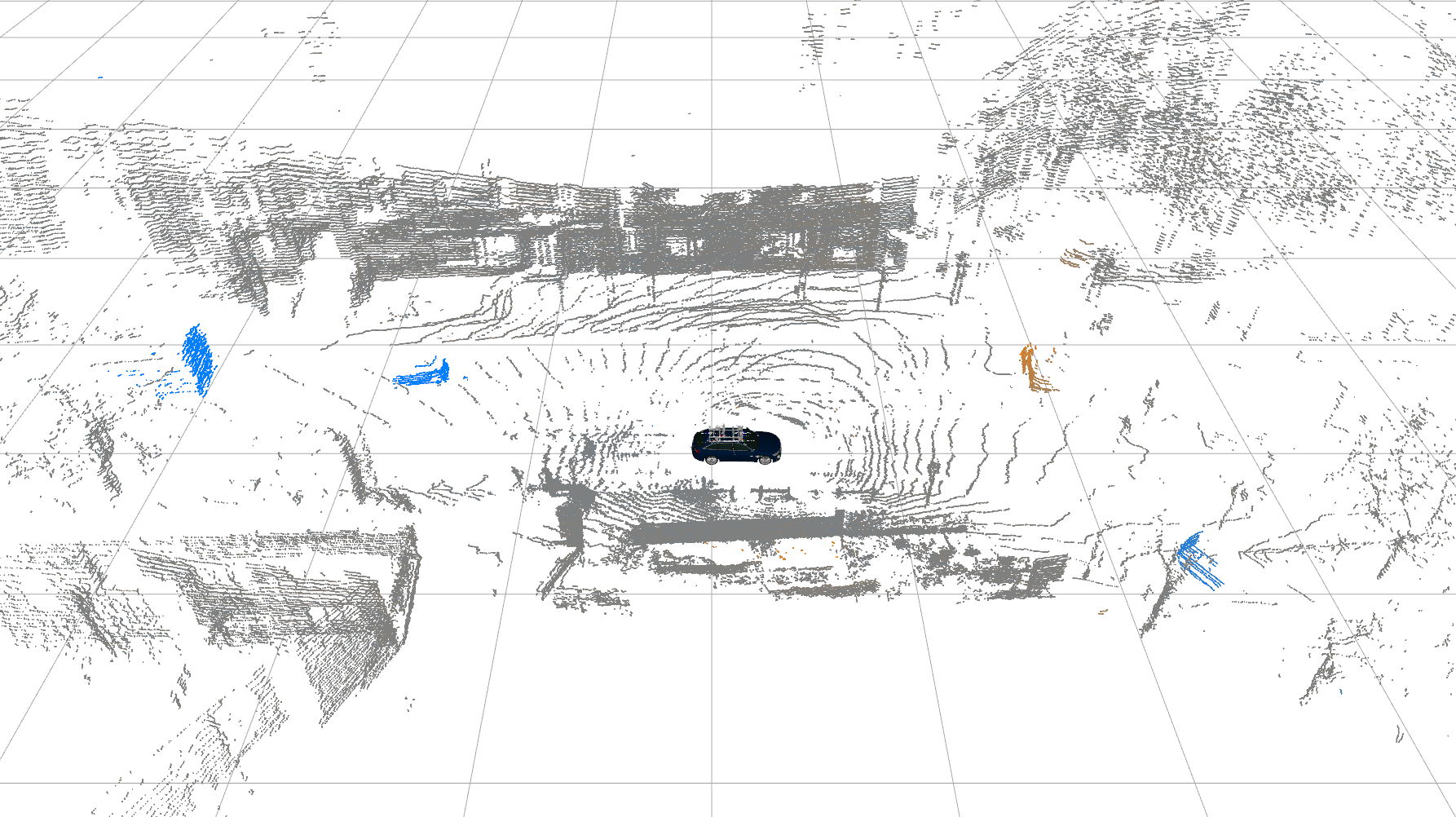}
        \caption{4D LiDAR}
        \label{fig:aeva}
    \end{subfigure}%
    \caption{Example point clouds of each LiDAR modality. \Cref{fig:os1} and \Cref{fig:os2} show the long and medium range \ang{360} LiDARs. The colors denote the reflectivity of the objects. \Cref{fig:aeva} depicts the 4D LiDAR sensors, where the colors represent the radial velocity of each point relative to the ego vehicle.}
\end{figure*}

The first group is used for object detection. To minimize vertical blind spots, it is integrated into the vehicle chassis. Therefore, the beams can be horizontal at the relevant height of our targets, not requiring a trade-off between close-range blind spots and range. To make use of the latest developments, we decided to mount 4D LiDARs based on frequency modulated continuous wave (FMCW) technology. These sensors allow the measurement of a radial velocity for each data point, therefore enhancing the available information. An illustrative example of the point cloud, including this velocity measurement, can be observed in \Cref{fig:aeva}. The group comprises of six AEVA Aeries II sensors. The main objectives pursued where a \ang{360} field of view (FOV) horizontally with minimal blind spots, as well as having sensors in the front of the vehicle facing laterally, helping in occluded intersections. Therefore, the sensors are positioned as follows: one sensor in the front of the vehicle, facing forward. Two sensors at the front corners, facing sideways. Two sensors mounted behind the front wheels, facing rearwards. One sensor in the back of the vehicle, facing backwards.
A feature of these sensors is the configurable FOV, ranging between \ang{120}, \ang{55} and \ang{35}. With decreasing FOV the resolution increase from 80 to 140 and 200 lines. Their maximum range of \SI{500}{\meter} further increases the benefits for perception tasks.

The second group consists of four \ang{360} Ouster OS1 LiDAR sensors mounted at the corners of the roof rack. This group serves as a reference setup for environmental perception. Having a redundant setup enables us to conduct various research such as cross-sensor and cross-domain experiments with data driven models. These sensors have a vertical FOV of \ang{45} with a state-of-the-art resolution of 128 lines. This high resolution coupled with their low noise predestines the sensors as reference setup.

The third group of sensors is used for LiDAR based SLAM localization. For this task, it is tolerated to overlook dynamic or potentially dynamic objects, which are often in near proximity of the car with limited height. Instead, the focus is on observing static targets such as buildings and lantern posts. Consequently, we positioned two Ouster OS2 sensors on an elevated position at the sides of our roof rack. The OS2 is a long range sensor with 128 lines, which uses a vertical FOV of \ang{22.5}.

\subsection{Cameras}

\begin{figure*}[t]
    \begin{subfigure}[b]{0.3\textwidth}
            \includegraphics[width=\textwidth]{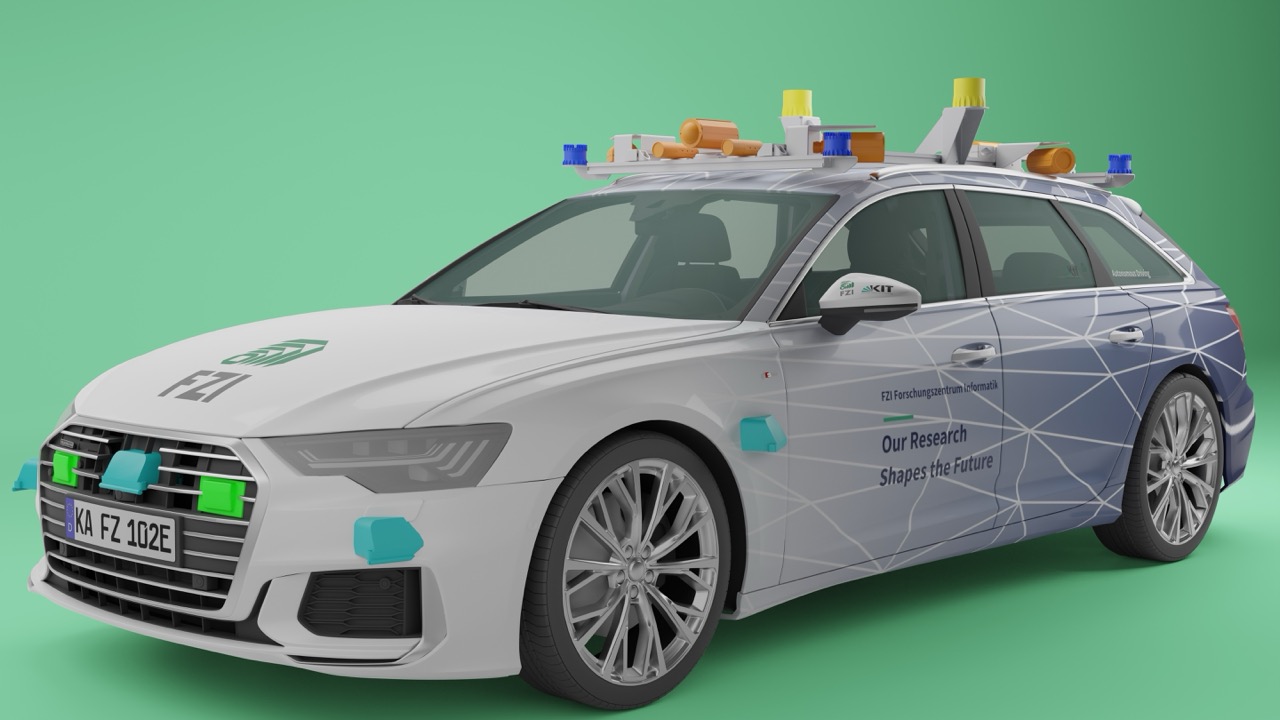}   
        \caption{The sensor positions.}
        \label{fig:cc-ng-rendering}
    \end{subfigure}%
    \hspace{0.02\textwidth}
    \begin{subfigure}[b]{0.3\textwidth}
            \includegraphics[width=\textwidth]{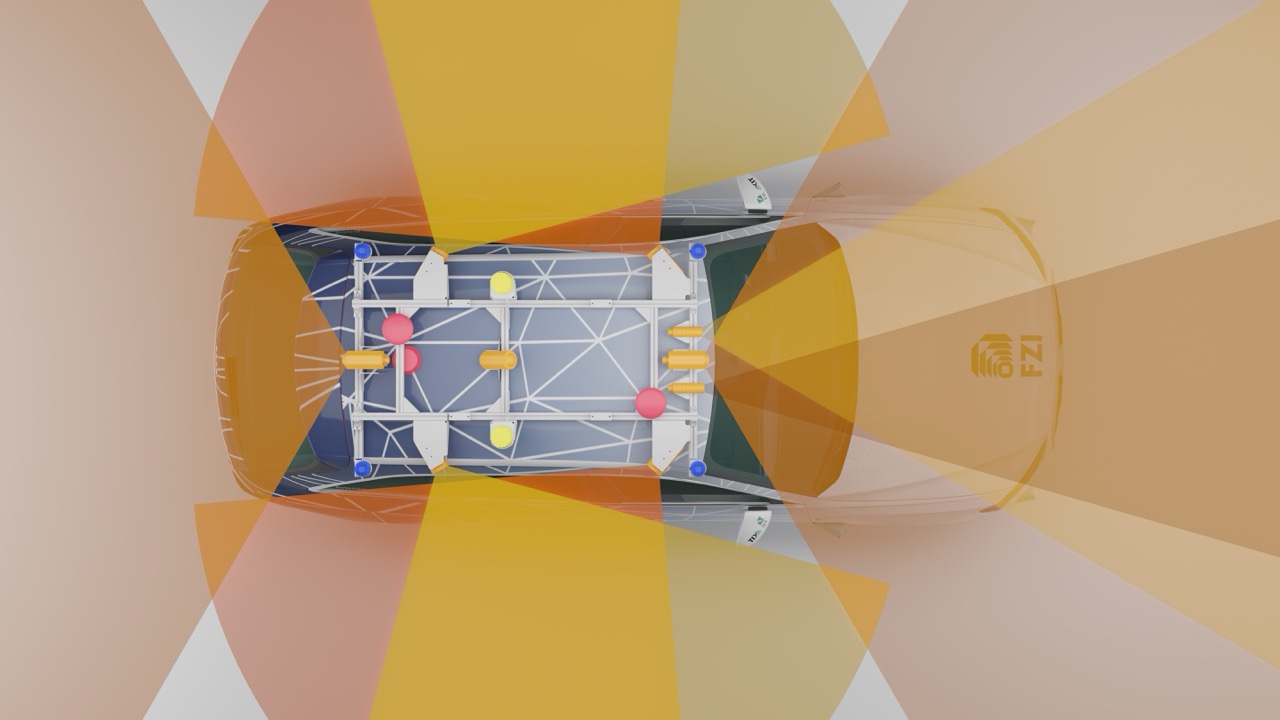}
            \caption{Camera positions and FOV.}
            \label{fig:camera_setup}
    \end{subfigure}%
    \hspace{0.02\textwidth}
    \begin{subfigure}[b]{0.3\textwidth}
            \includegraphics[width=\columnwidth]{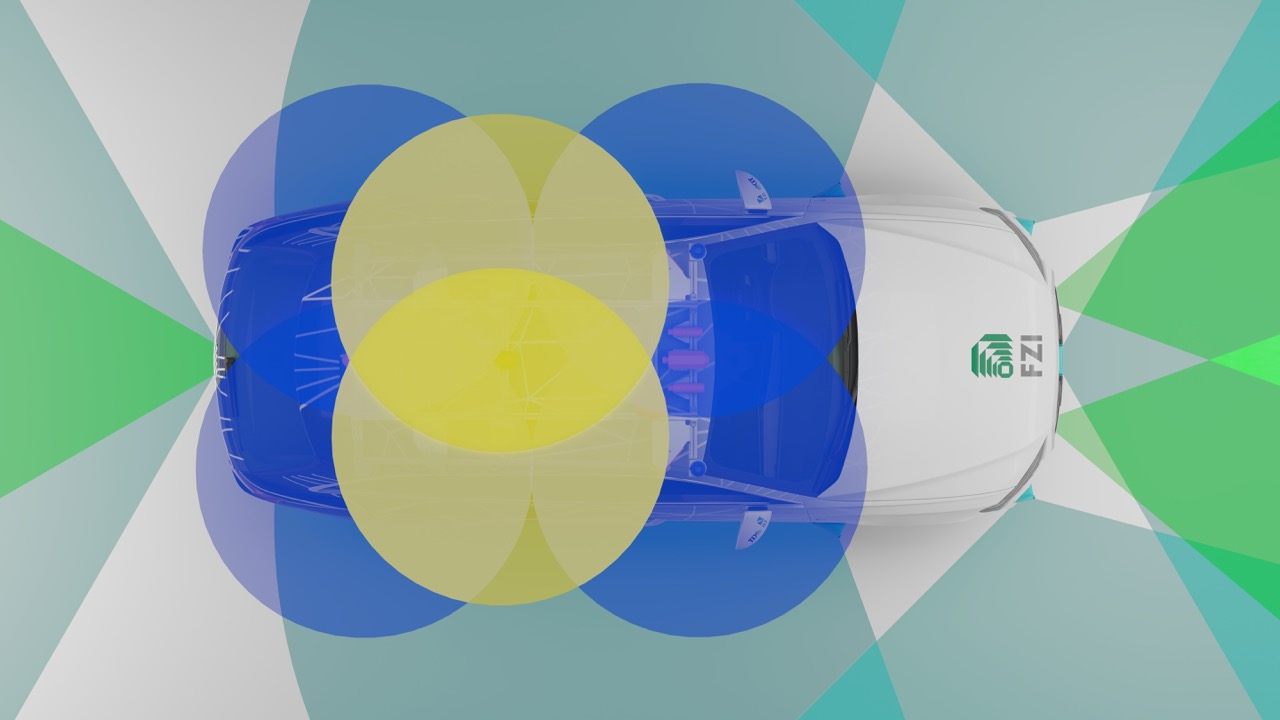}
            \caption{LiDAR and radar positions and FOV.}
            \label{fig:sensor_setup}
    \end{subfigure}%
\caption{Sensor setup of CoCar NextGen. 4D-LiDAR (cyan), 360°-LiDAR (blue close-range, yellow far-range), Full-HD cameras (orange), Radar (green)}
\end{figure*}

The goal was again to create a setup with \ang{360} FOV and small blind spots.
Firstly, we use three cameras with different focal lengths facing forward, to be able to have both a wide FOV as well as good resolution at medium to long distances.
Additionally, we have two cameras with a cross-eyed setup on each side of the vehicle, one mounted in the front, facing rearward, and one further back, facing forward. This reduces blind spots significantly.
Another camera is mounted at the rear to complete the \ang{360} FOV. Lastly, we have an additional camera in the middle of the roof looking upwards, as seen in \Cref{fig:camera_setup}.
As traffic lights in Germany are often located right above the stop line, they might otherwise not be visible in the front wide camera..
The camera parameters can be seen in \Cref{tab:camera_fovs}.
Another objective was to position the cameras in proximity to the \ang{360} LiDARs. Therefore, parallax errors are minimized when projecting LiDAR point clouds into the camera image. This simplifies data labeling, as well as  the fusion of semantic information from the camera with 3D information of the LiDAR sensors.

\begin{table}
\centering
\resizebox{\linewidth}{!}{%
\begin{tabular}{lcccc} 
 \toprule
 \multirow{2}{*}{Camera} & \multirow{2}{*}{Type}  & FOV& Resolution & FPS\\
  & &[°] & [px] & [Hz]\\
 \midrule
 Front Wide & acA2500-20gc & $120\times90$ & 2590$\times$2048 & 20\\ 
 Front Medium& a2A1920-51gcPRO & $90\times55$ & 1920$\times$1200 & 50\\
 Front Tele &a2A1920-51gcPRO & $32\times20$ & 1920$\times$1200 & 50\\
 Side Forwards (x2)& acA1920-40gc & $120\times90$ & 1920$\times$1200 & 40\\
 Side Rearwards (x2)& acA1920-40gc & $120\times90$ & 1920$\times$1200 & 40\\
 Rear Wide & acA2500-20gc & $120\times90$ & 2590$\times$2048 & 20\\
 Traffic Light (roof)& acA2500-20gc & $120\times90$ & 2590$\times$2048 & 20\\
 \bottomrule
\end{tabular}
}
\caption{Camera field of views, resolutions and frequencies}
\label{tab:camera_fovs}
\end{table}

\begin{figure}[t]
    \newsavebox{\largesttop}%
    \newsavebox{\largestmid}%
    \newsavebox{\largestbot}%
    \savebox{\largesttop}{\includegraphics[width=0.32\columnwidth]{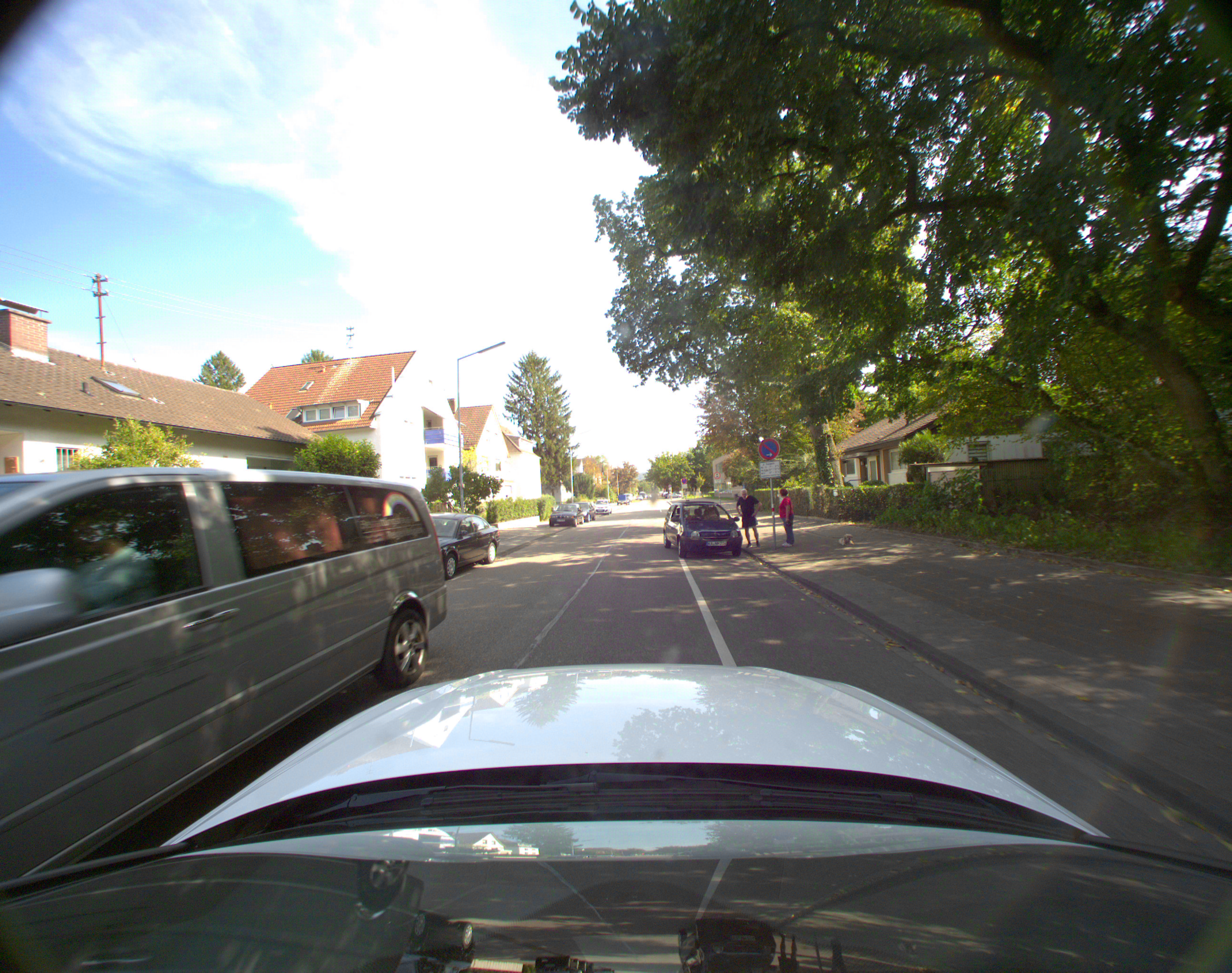}}%
    \savebox{\largestmid}{\includegraphics[width=0.32\columnwidth]{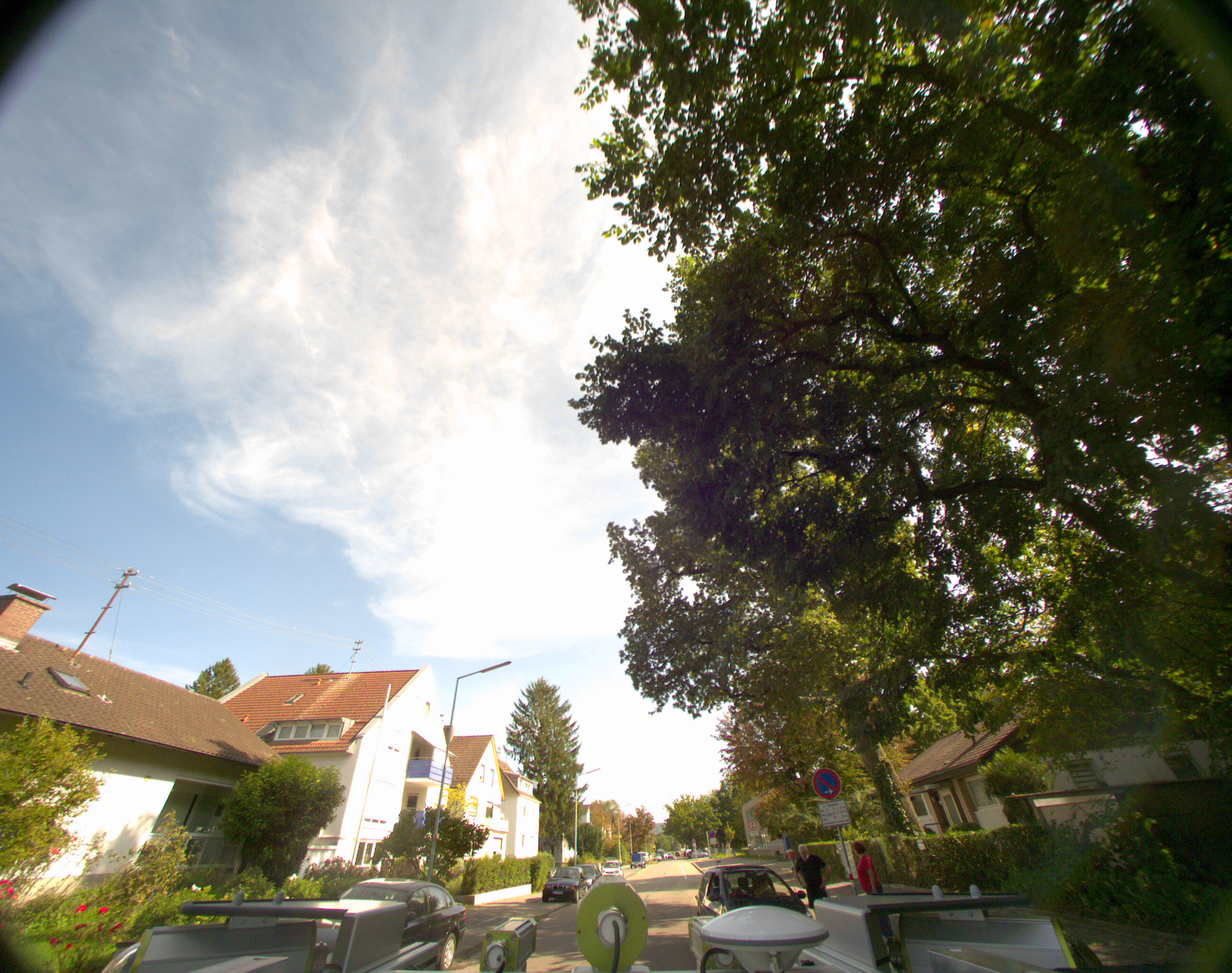}}%
    \savebox{\largestbot}{\includegraphics[width=0.32\columnwidth]{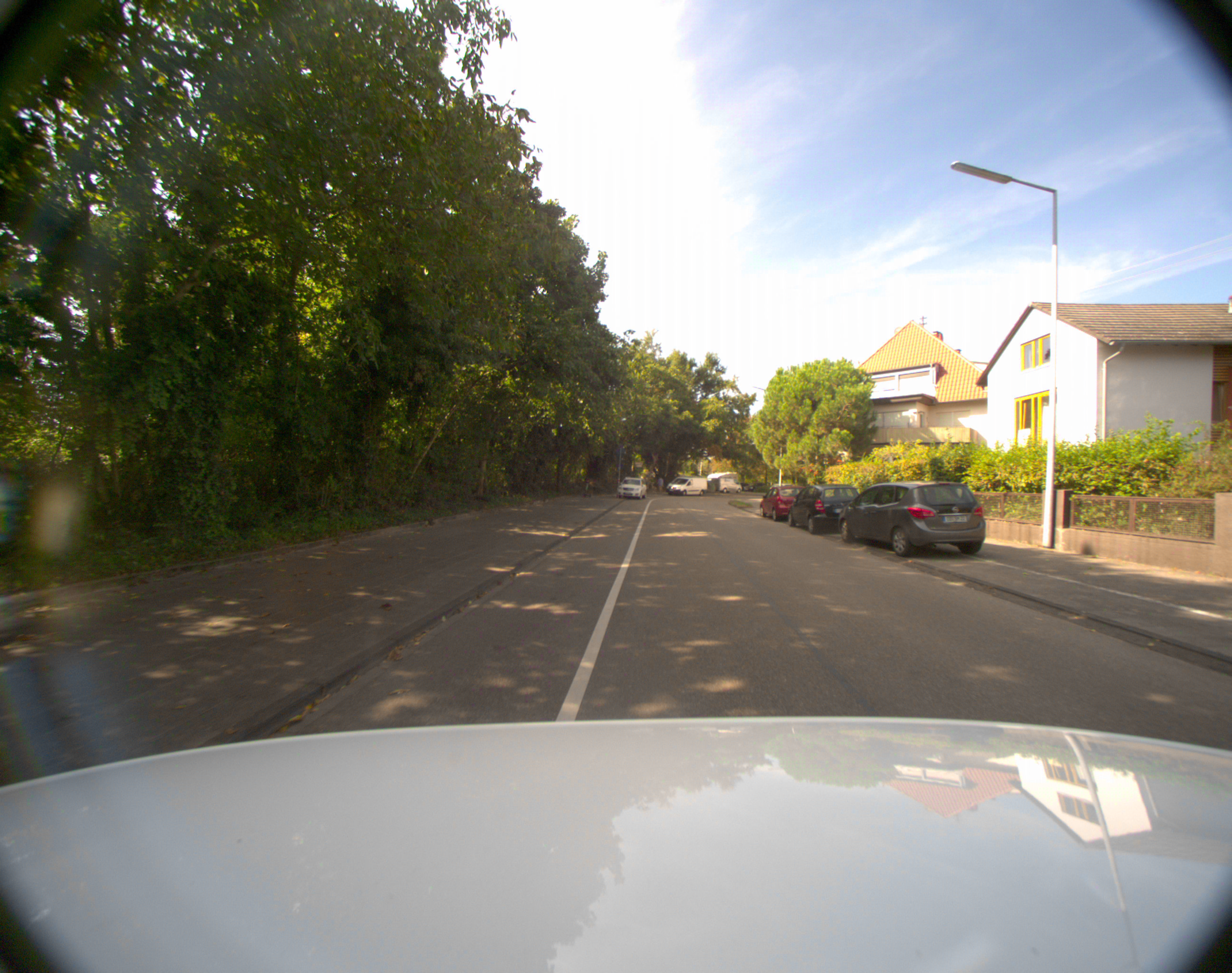}}%
    \begin{subfigure}[t]{0.32\columnwidth}
        \raisebox{\dimexpr.5\ht\largesttop-.5\height}{%
        \includegraphics[width=\textwidth]{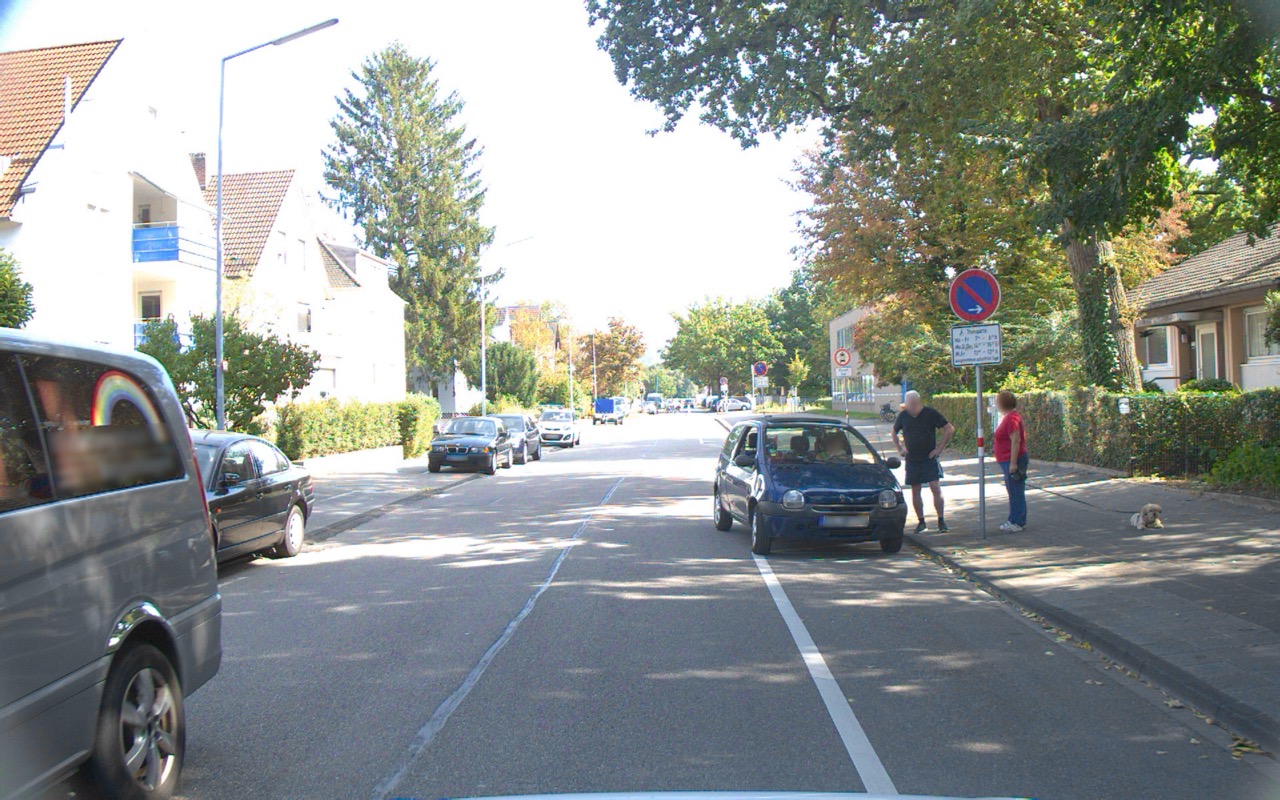}%
        }
        \caption{Front Medium}
        \label{fig:front}
    \end{subfigure}%
    \hfill%
    \begin{subfigure}[t]{0.32\columnwidth}
        \includegraphics[width=\textwidth]{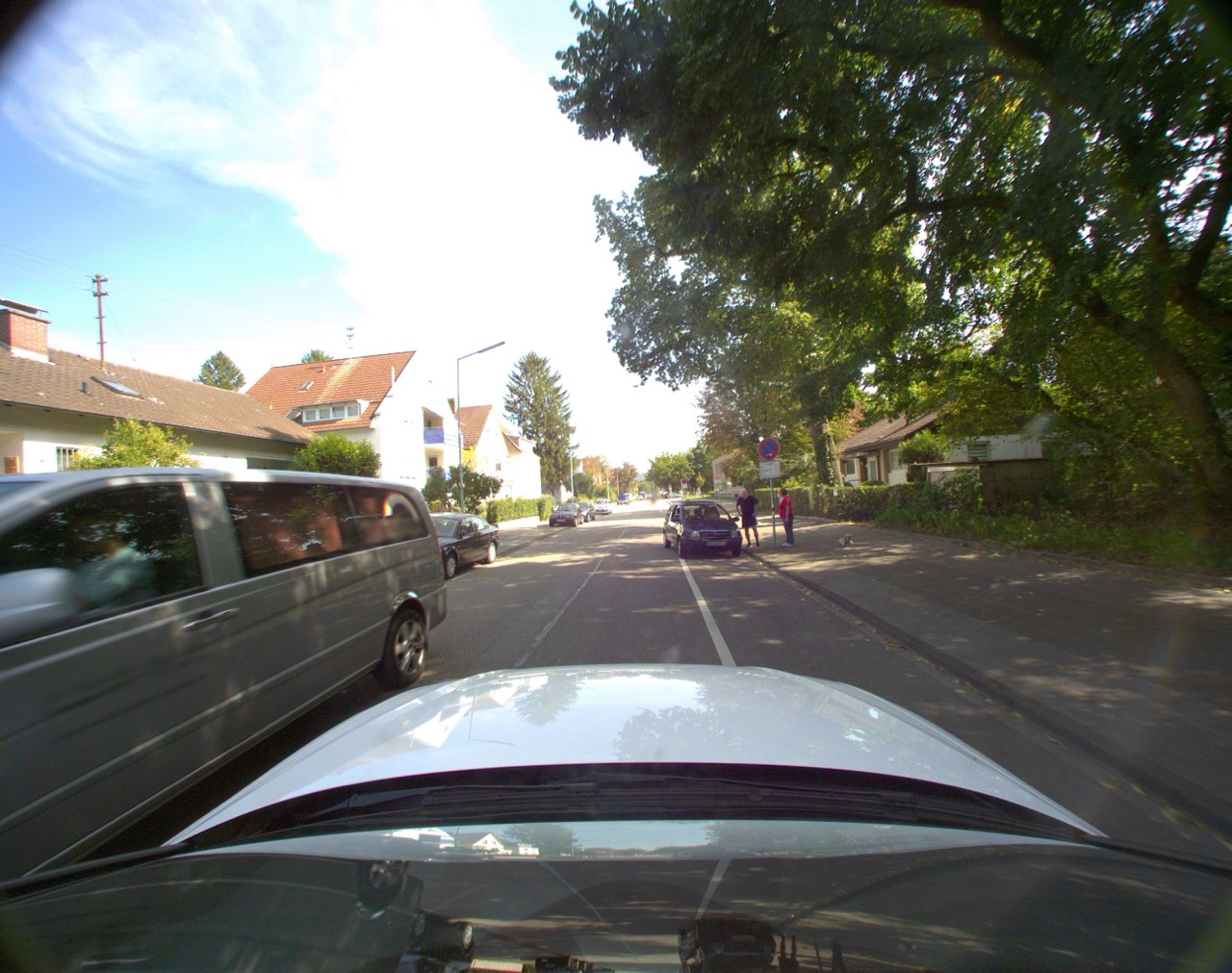}
        \caption{Front Wide}
        \label{fig:front_wide}
    \end{subfigure}%
    \hfill%
    \begin{subfigure}[t]{0.32\columnwidth}
        \raisebox{\dimexpr.5\ht\largesttop-.5\height}{%
        \includegraphics[width=\textwidth]{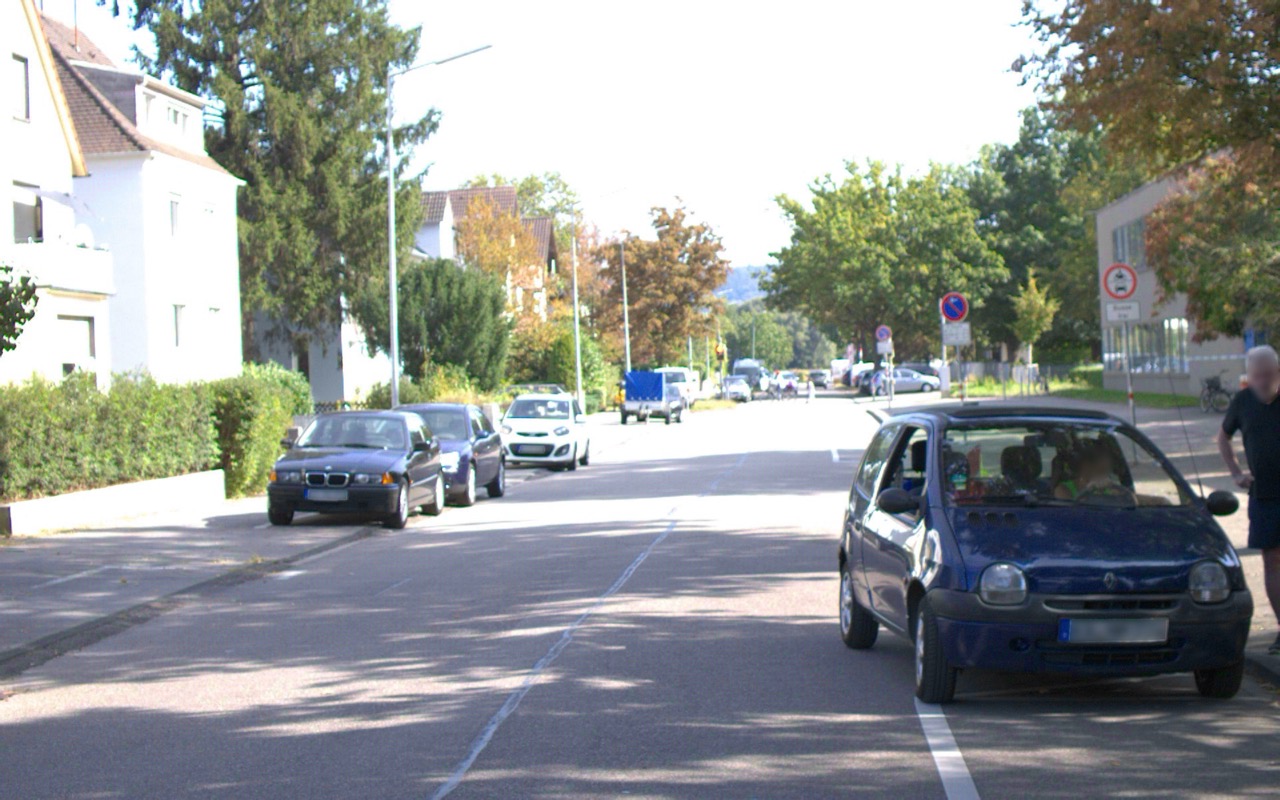}%
        }
        \caption{Front Tele}
        \label{fig:front_tele}
    \end{subfigure}\vspace{0.02\linewidth}
    \begin{subfigure}[t]{0.32\columnwidth}
         \raisebox{\dimexpr.5\ht\largestmid-.5\height}{%
        \includegraphics[width=\textwidth]{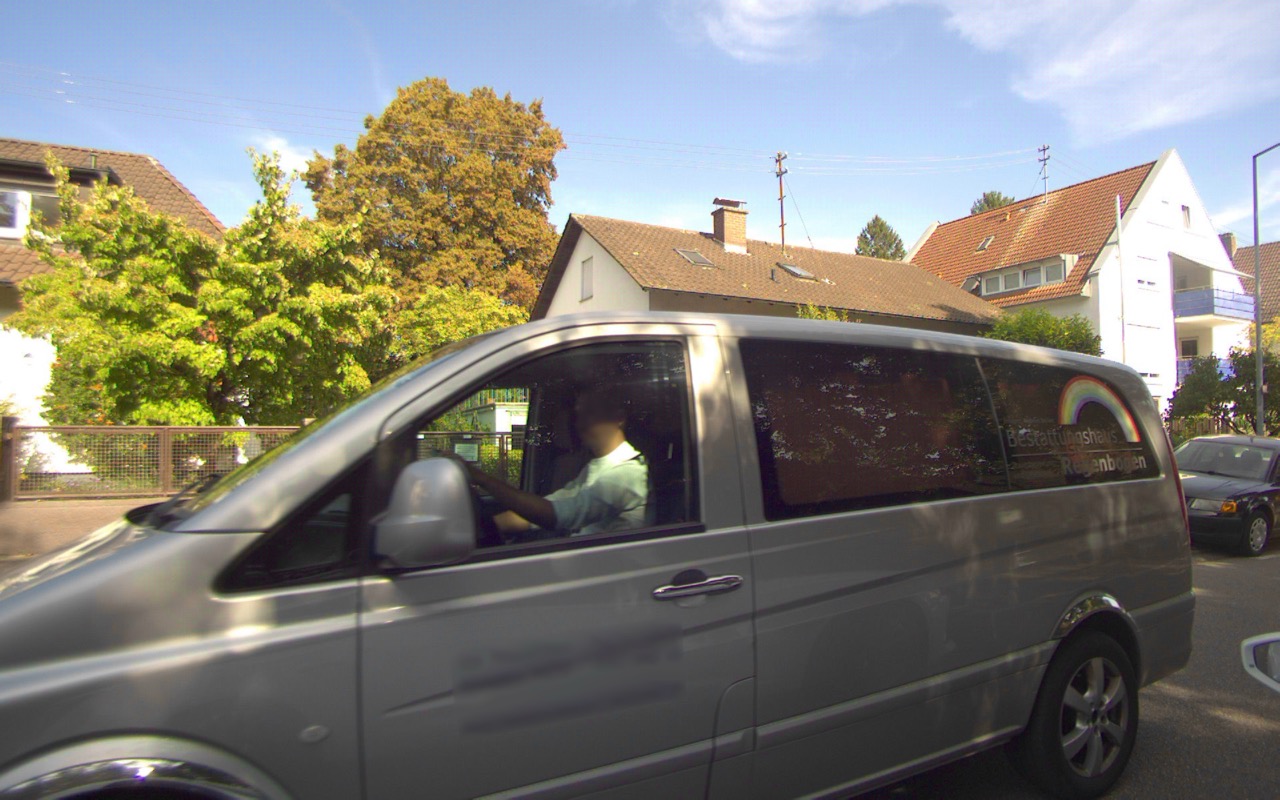}%
        }
        \caption{Left Forward}
        \label{fig:left_forward}
    \end{subfigure}%
    \hfill%
    \begin{subfigure}[t]{0.32\columnwidth}
        \includegraphics[width=\textwidth]{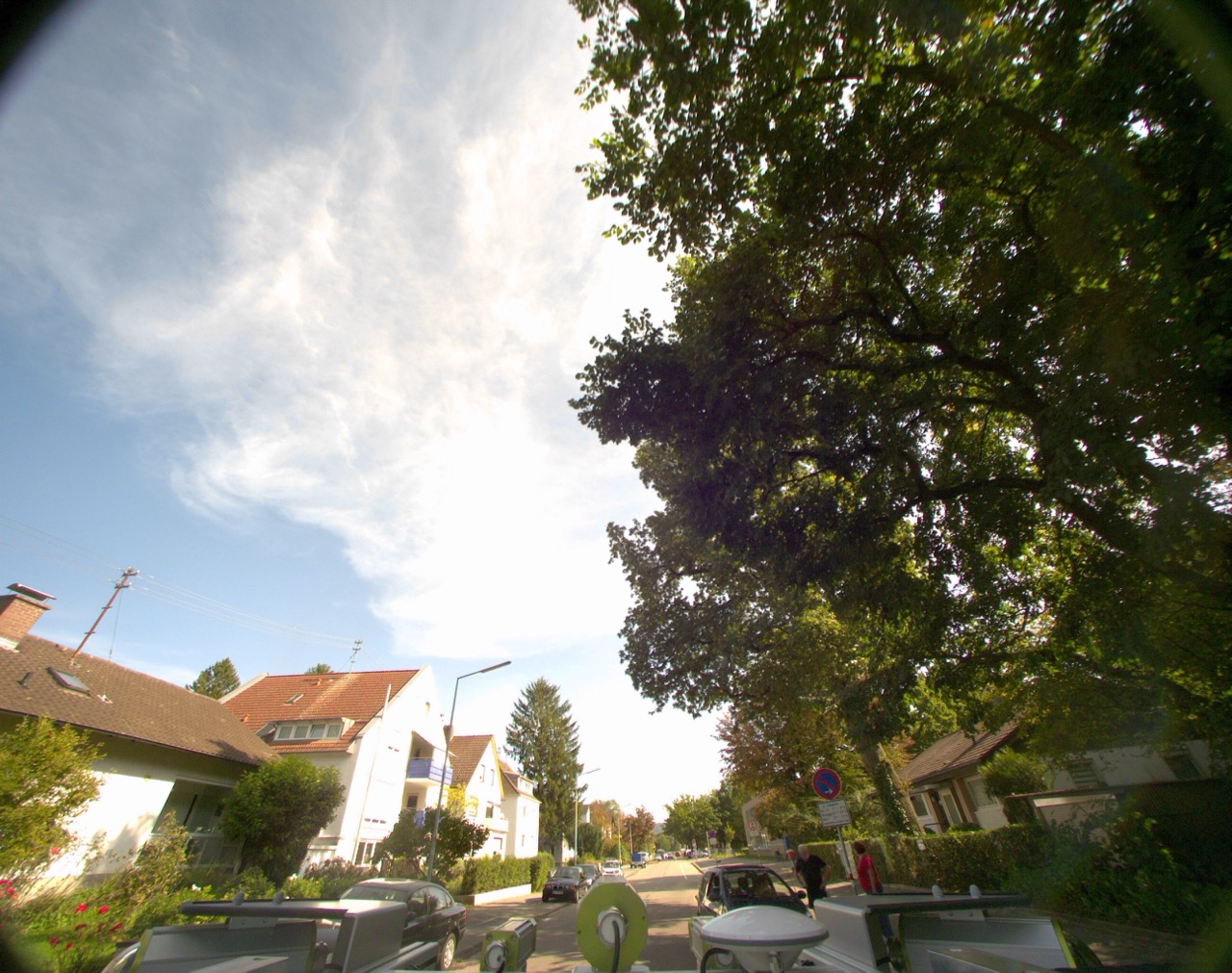}
        \caption{Traffic-Light}
        \label{fig:traffic light}
    \end{subfigure}%
    \hfill%
     \begin{subfigure}[t]{0.32\columnwidth}
		\raisebox{\dimexpr.5\ht\largestmid-.5\height}{%
        \includegraphics[width=\textwidth]{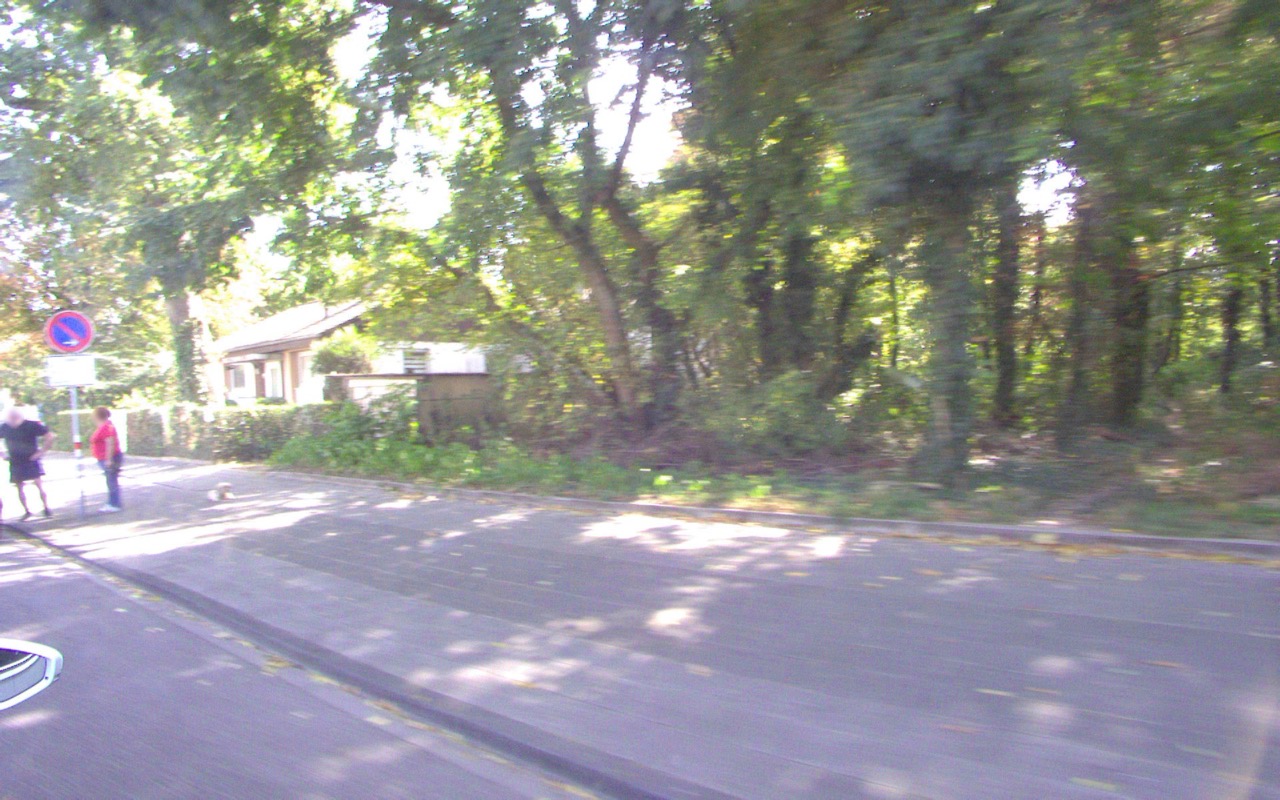}%
        }
        \caption{Right Forward}
		\label{fig:right_forward}
    \end{subfigure}\vspace{0.02\linewidth}
    \begin{subfigure}[t]{0.32\columnwidth}
        \raisebox{\dimexpr.5\ht\largestbot-.5\height}{%
        \includegraphics[width=\textwidth]{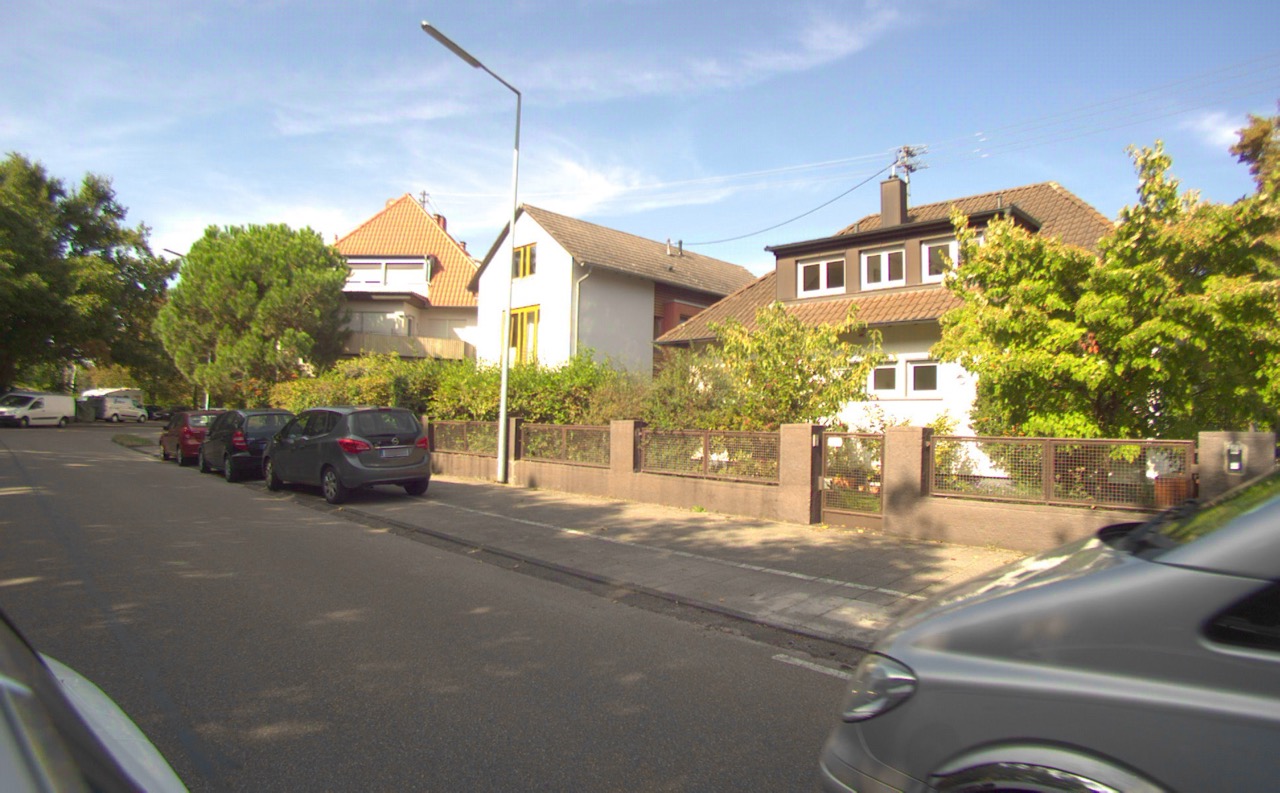}%
        }
        \caption{Left Rearward}
        \label{fig:left_reward}
    \end{subfigure}%
    \hfill%
    \begin{subfigure}[t]{0.32\columnwidth}
        \includegraphics[width=\textwidth]{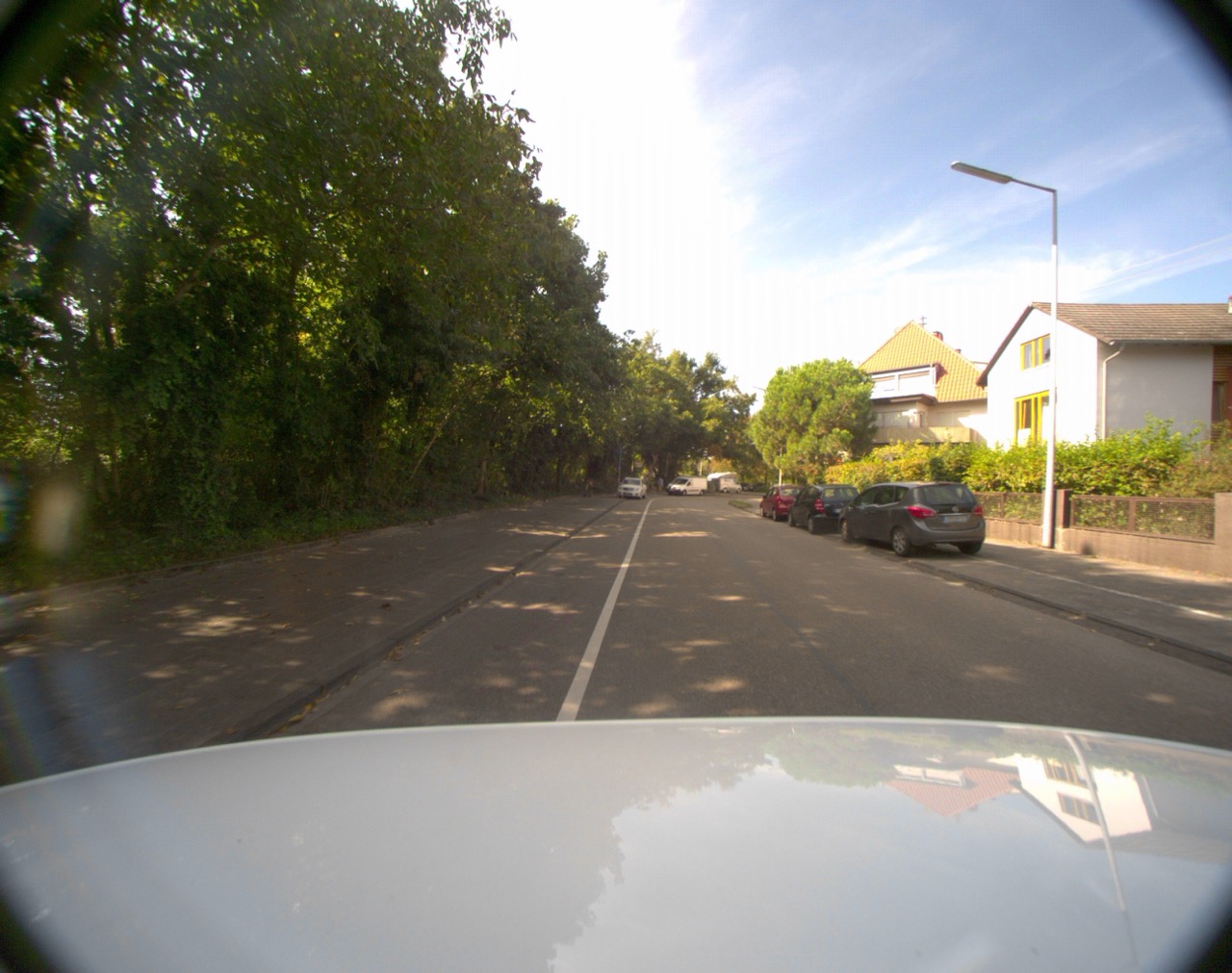}
        \caption{Rear Wide}
        \label{fig:rear_wide}
    \end{subfigure}%
    \hfill%
    \begin{subfigure}[t]{0.32\columnwidth}
        \raisebox{\dimexpr.5\ht\largestbot-.5\height}{%
        \includegraphics[width=\textwidth]{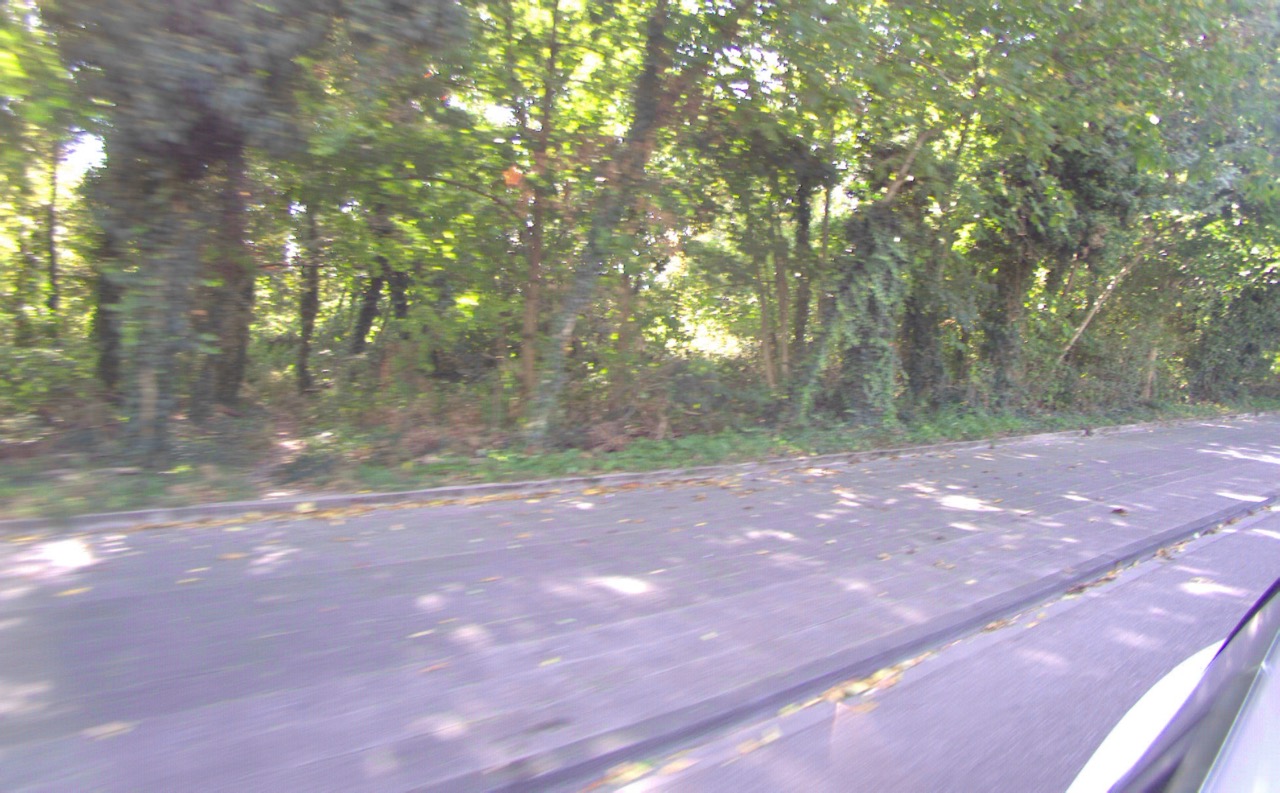}%
        }
        \caption{Right Rearw.}
        \label{fig:right_reward}
    \end{subfigure}%
    \caption{Example images of the nine cameras built onto the rooftop structure.}
    \label{fig:camera_example}
\end{figure}

\subsection{Radar Sensors}

To achieve comprehensive sensor measurements, we have augmented our sensor suite with three Continental AR548 RDI radar sensors. These state-of-the-art sensors feature 192 antennas and advanced 4D capabilities with a configurable horizontal FOV of up to \ang{120}. For scenarios involving higher speeds, the sensors are strategically positioned, with two located at the front and one at the rear.

\subsection{Wireless Communication}
Our research approach includes communication with other intelligent vehicles and connected infrastructure (V2X). Therefore, CoCar NextGen is equipped with LTE, 5G as well as Car2X 802.11p WiFi antennas. To facilitate demonstrations and enhance user-friendliness, a WiFi network is established. This network enables wireless access to all components in the vehicle network. 

\subsection{Human-Machine Interface}

\begin{figure*}[t]
    \begin{subfigure}[t]{0.15\textwidth}
        \includegraphics[height=3.5cm]{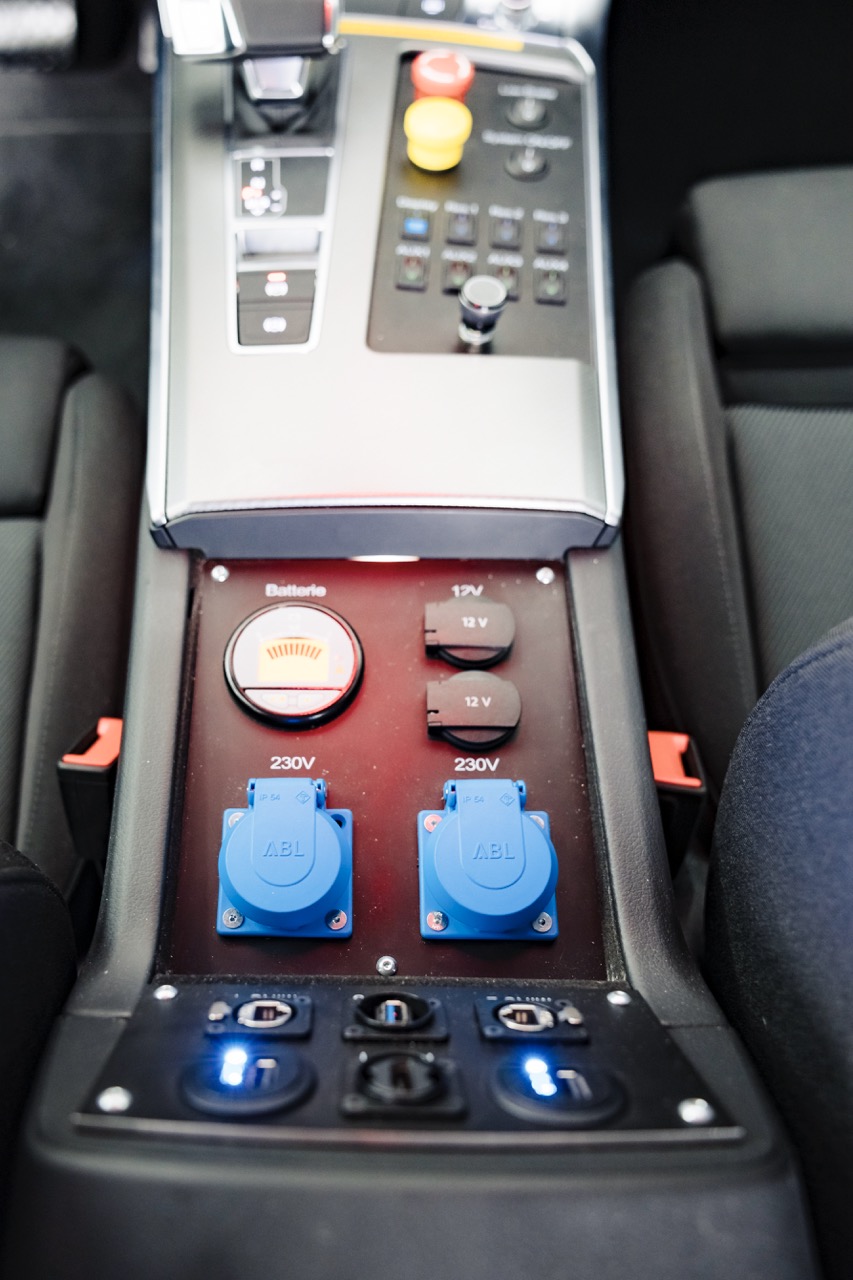}
        \caption{Center console}
        \label{fig:hmi-inputs}
    \end{subfigure}%
    \hspace{0.02\textwidth}
    \begin{subfigure}[t]{0.15\textwidth}
        \includegraphics[height=3.5cm]{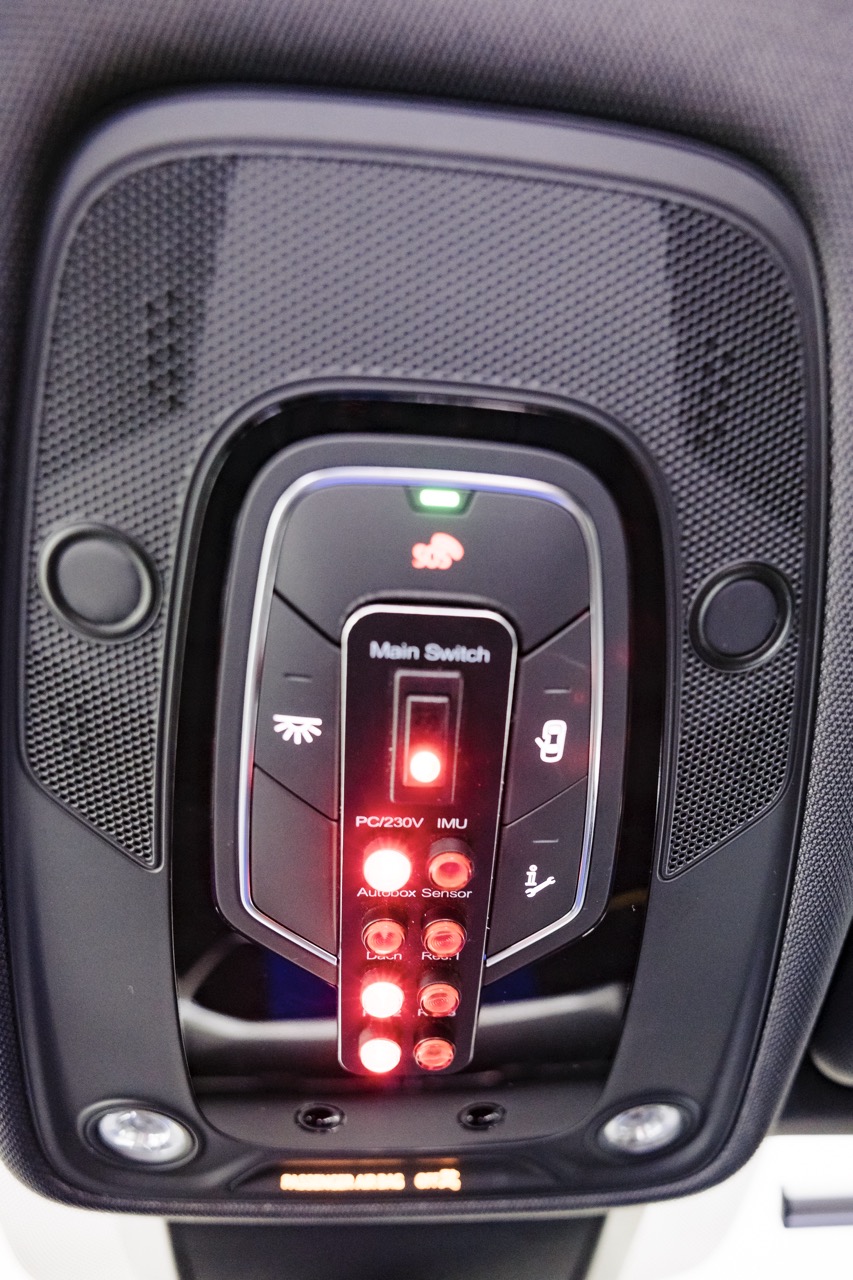}
        \caption{Headliner}
        \label{fig:switches}
    \end{subfigure}%
    \hspace{0.02\textwidth}
    \begin{subfigure}[t]{0.3\textwidth}
        \includegraphics[height=3.5cm]{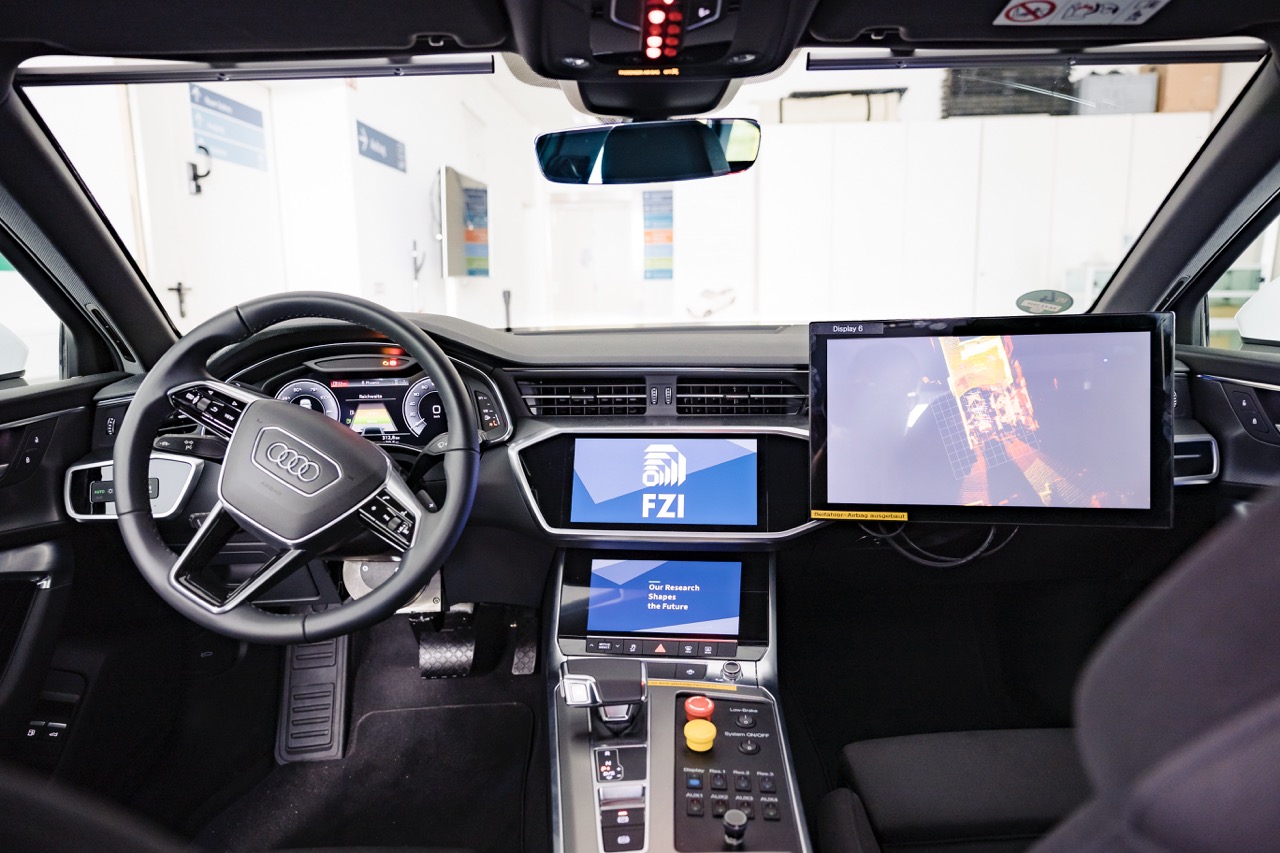}
        \caption{Cockpit}
        \label{fig:front-displays}
    \end{subfigure}%
    \hspace{0.02\textwidth}
    \begin{subfigure}[t]{0.3\textwidth}
        \centering
        \includegraphics[height=3.5cm]{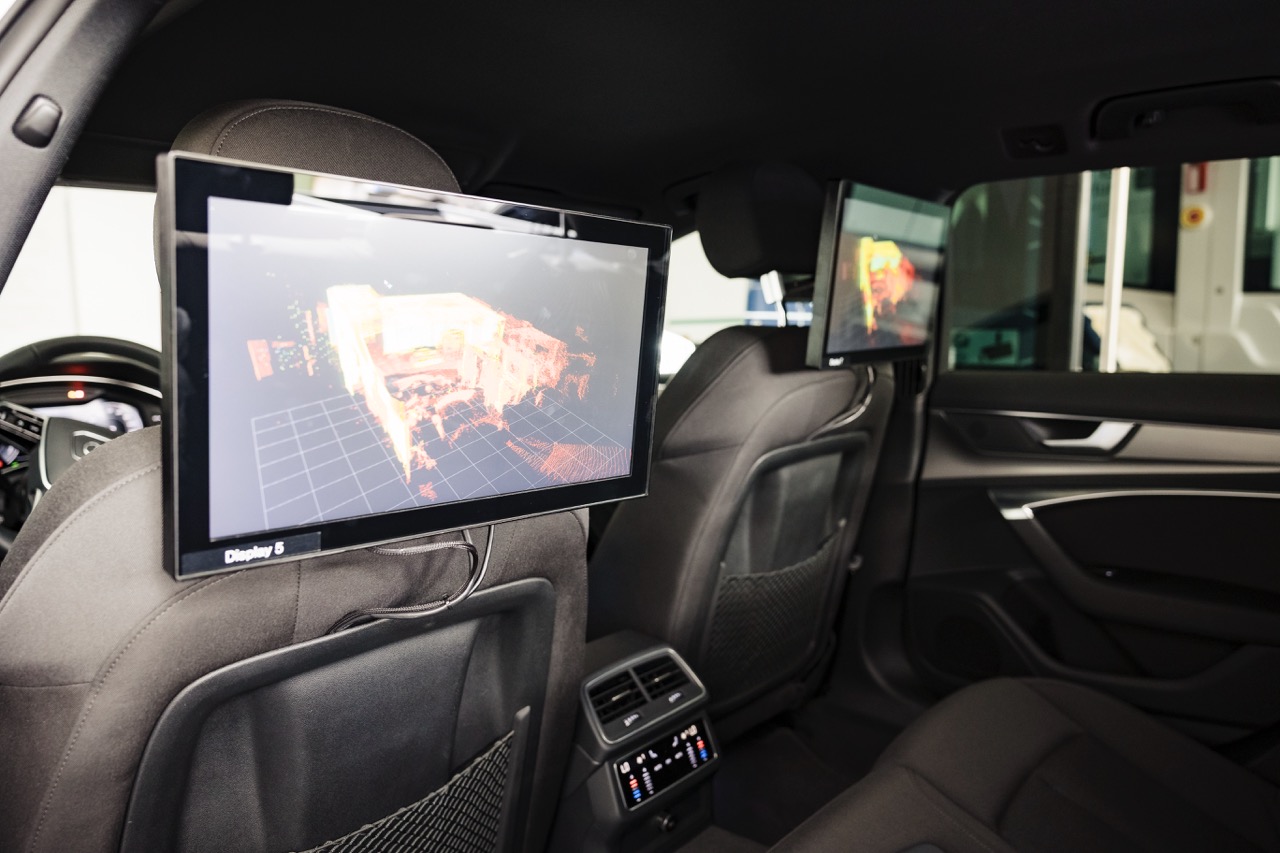}
        \caption{Rear seats}
        \label{fig:rear-displays}
    \end{subfigure}%
    \caption{Human-Machine-Interfaces of CoCar NextGen. Our aim was to maximise usability and flexibility. The center console contains the interfaces for the drive-by-wire system as well as programmable hardware switches. Moreover, there are power outlets, the battery display and some connectivity ports (RJ45, USB, HDMI). The headliner console contains the buttons to switch the power of each component group. In the cockpit there is a passenger display as well as the infotainment display, which can be fed with custom content. In the rear there are two displays behind the front headrests.}
\end{figure*}

CoCar NextGen is envisioned not solely as a sensor platform but also as a versatile demonstration vehicle and research platform. Recognizing this multifaceted role, the Human-Machine Interface (HMI) is designed to be adaptable and capable of accommodating evolving requirements. The HMI encompasses two primary aspects: the presentation of information and the interface for automated driving functions. To achieve the former, the vehicle incorporates four displays, including two for the rear seat occupants and one for the front passenger seat. Furthermore, supplementary information can be presented on the vehicle's infotainment displays, as depicted in \Cref{fig:front-displays,fig:rear-displays}. The displays can also be fed by external sources via HDMI ports in the interior. Those ports can also be used to display vehicle information on external displays.
To connect to the compute platform and various other components, multiple network sockets (RJ45) and USB ports are placed in the interior. The interfaces are located within the glovebox and the center console. The latter can be seen in \Cref{fig:hmi-inputs}.
Furthermore, programmable buttons have been integrated into the car's headlining. These switches are configured to control the power supply to distinct sensor modalities, enabling individual activation as needed. The switches can be seen in \Cref{fig:switches}. Moreover, additional programmable switches and a incremental encoder are placed in the center console. These are connected to the compute platform via a GPIO interface.
Finally, the center console contains buttons for the drive-by-wire system as well as a display for the state of charge of the component battery, see \Cref{ssec:power}.

\subsection{Computing Platform} %
\label{ssec:compute}
\begin{figure}[t]
    \includegraphics[width=\columnwidth]{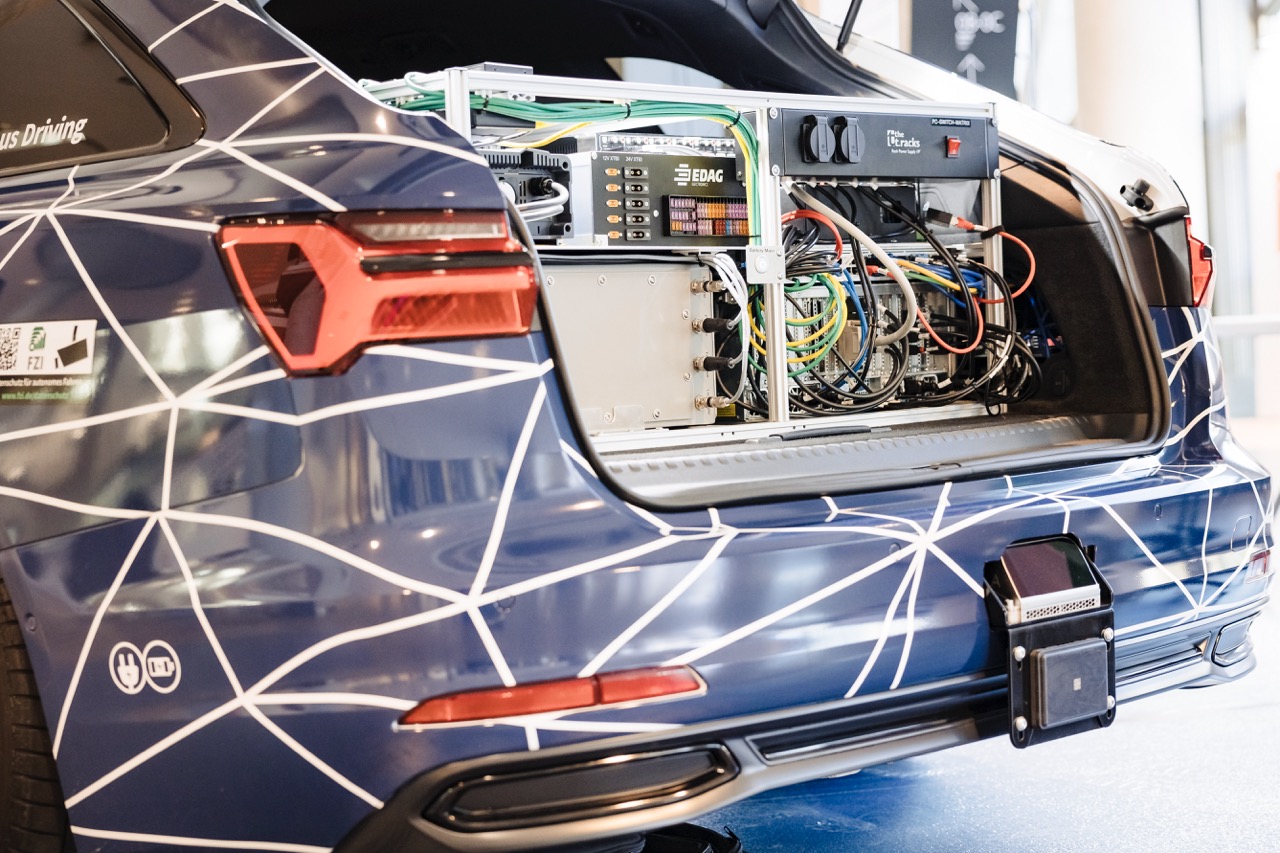}
    \caption{Trunk of CoCar NextGen with the network switch and the computing platform visible on the right-hand side. In the lower left, the additional battery is visible in grey.}
    \label{fig:cc-ng-rear}
\end{figure}
To process the sensor data and to run advanced automated driving software, a high performance computing platform is installed.
As we intend to support a variety of use cases, the flexibility of our computing platform is of utmost importance.
Therefore, we choose a standard general purpose server running a Linux OS rather than an embedded system, or specialized single-purpose accelerators.
This liberates us from the need to adapt our software for specialized hardware.

In contrast to the traditional automotive network hierarchy of using multiple smaller components for each zone or domain,~\cite{bello2018recent} we employ a single centralized compute server. This machine aggregates and processes all sensor data in one location.
The main reason for this centralization is a reduction in latency. Moreover, this offers a large improvement in bandwidth between different software components. To facilitate parallel operation without competing for resources, our platform offers a high thread-count.

The main server is equipped with two Intel Xeon Platinum 8352M 32-core CPUs running at \SI{2.3}{\giga\hertz}, for a total of 128 threads, supported by \SI{768}{\giga\byte} of RAM.
Three Nvidia A6000 GPUs with \SI{38.7}{TFLOPS} and \SI{48}{\giga\byte} VRAM each are integrated to provide acceleration for deep-learning based components and other highly parallel tasks.
Eight \SI{7.68}{\tera\byte} NVME-SSDs in a RAID-0 configuration are used for data recording. With this arrangement, we achieve a total measured disk write bandwidth of approximately \SI{33.4}{\giga\bit/\second}. This is sufficient for the total bandwidth of \SI{20}{\giga\bit/\second} of all sensors combined, providing some extra headroom for inter process communication.
The operating system, as well as all software, is stored on a RAID-1 of two \SI{960}{\giga\byte} SSDs for redundancy and reliability.
Network interfacing is provided by four SFP+ ports with \SI{10}{\giga\bit/\second} bandwidth each.
Additional interfaces include a vehicle CAN card with four DSUB9 ports, two USB 3.1 cards as well as a GPIO card with an Analog-to-Digital converter.

To optimize performance, the PCIE-Lane assignment was optimized. All storage drives were assigned to one CPU, while networking and the GPUs were assigned to the other. This aligns with the typical data-flow of the driving functions: Raw sensor data is received from the network, and processed by compute-intensive workloads such as perception tasks. Afterwards, lower bandwidth information of higher level of abstraction can be sent across the CPU interconnect for later processing stages, such as trajectory planning. Nonetheless, the CPU interconnect has enough bandwidth to transport the raw data between the CPU sockets, such as when recording raw sensor data to disk.
An auxiliary real-time embedded dSPACE MicroAutobox III is used for real-time control logic for the vehicle actuators. Against our aim for a single general purpose solution, this design was chosen to provide robustness for the task of motion control.

\subsection{Network Configuration} %
As a modern research vehicle with a multitude of sensors and network-based peripherals, our vehicle produces a large bandwidth of sensor data, which needs to be gathered and processed with low latency.
Under the conservative assumption that all sensors and other peripherals can saturate their network link, the vehicle network needs to provide at least \SI{20}{\giga\bit/\second} in network bandwidth.
With few exceptions, the communication path for our vehicle network is mainly a many-to-one architecture, with the compute server at the center, and connections fanning in from the individual sensors and peripherals.
Therefore, we aggregate the the compute server's four SFP+ Ethernet ports to a central network switch via a via an IEEE 802.3ad bond. The aggregation provides a bandwidth reaching \SI{40}{\giga\bit/\second}.
The switch is then connected to the individual components via individual RJ45 \SI{1}{\giga\bit/\second} Ethernet ports.
We chose a managed 52-Port 1-Gigabit Ethernet switch with 4 SFP+ 10-Gigabit ports for this purpose, which also provides Power-over-Ethernet (PoE) for our PoE-based cameras.
Where necessary, components with Automotive Ethernet connections (1000-Base-T1) are adapted via appropriate media converters to regular consumer Ethernet (1000-Base-T) before being connected to the central switch.

To reduce network traffic, as well as to inhibit interference between unrelated components, the different components are grouped into VLAN networks by component type.
For instance, the sensors modalities each are separated in their own VLAN network.
Through the use of tagged Ethernet ports on the network switch, the VLAN networks are transparent to the components. These only observe their own virtual network, composed of their peers, as well as the compute server. The latter receives tagged packets through an aggregated link and can thereby communicate with all networks.
The VLAN architecture avoids the latency and bandwidth overhead of complex routing, while also being simpler to configure. Therefore, this setup is more reliable and space-efficient than using multiple discrete networking devices.

The vehicle has an integrated IoT Router, which provides internet access via a 5G mobile network connection, or a \SI{1}{\giga\bit/\second} ethernet uplink.
The router automatically switches between different uplinks based on availability, and thus provides a reliable internet connection to the server, as well as the GNSS module for DGPS correction data.

In many applications, having synchronized, or accurately time-stamped data between different sensors is vital.
For an accurate time synchronization between all time-sensitive components, mainly the sensors, the vehicle server provides a PTP clock synchronization to the relevant VLAN networks.
To synchronize the vehicle with external infrastructure, the GPS timestamp from the GNSS module is used as the on-board grand-master clock.

\subsection{Vehicle Interface}
\label{ssec:vehicle-interface}
To enable closed-loop automated driving, a drive-by-wire system is installed. To maintain safety and road approval, the steering system is equipped with a magnetic clutch with \SI{11}{\newton\meter} of torque. The pedals are operated with mechanical actuators, allowing manual override at any time. Additionally, the system can operate secondary functions such as indicators, headlights, horn, windscreen wipers and the gearbox via CAN.
The system is controlled via a real-time embedded compute unit connected to the drive-by-wire system via can.

\subsection{Power-Supply}
\label{ssec:power}

\begin{figure}[ht]
	\centering
		\includegraphics[width=\columnwidth]{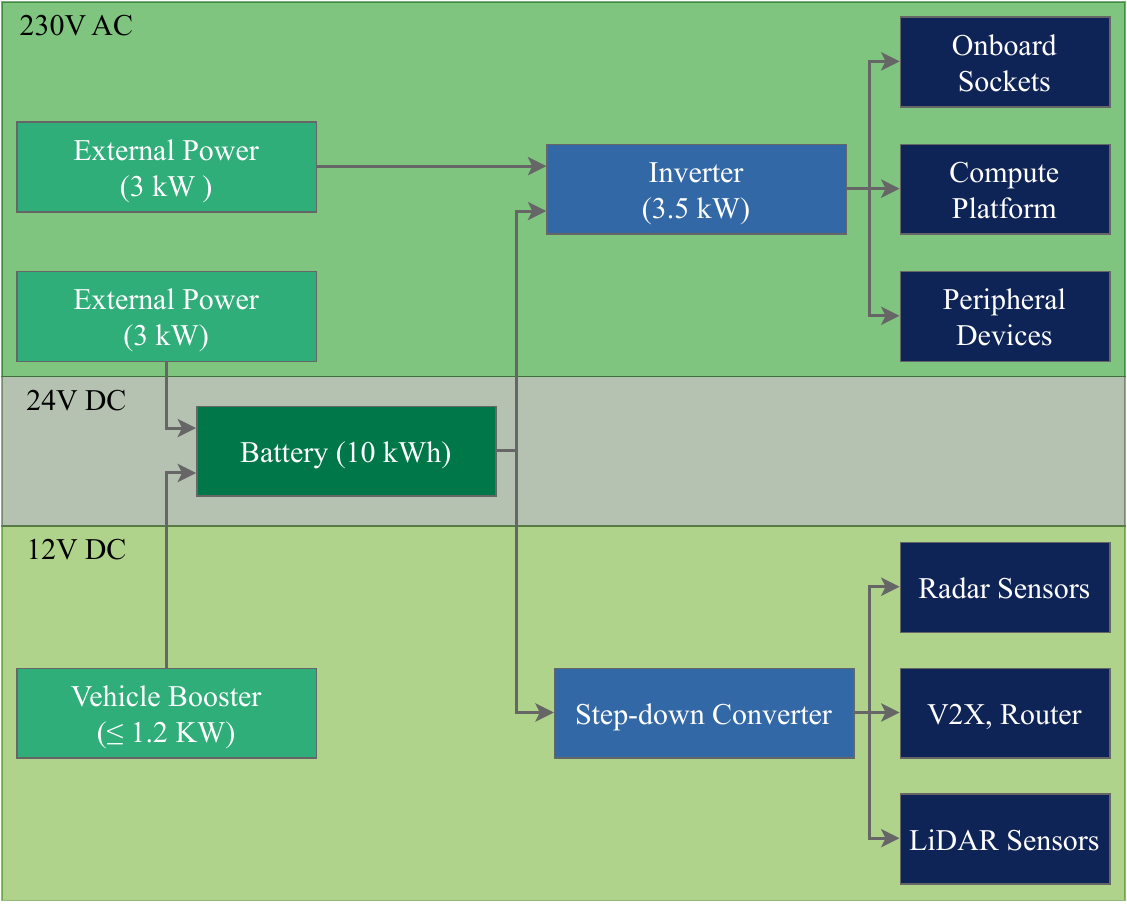}
		\caption{An overview of the on-board power system. The system is powered through a \SI{24}{\volt} DC battery. The \SI{24}{\volt} DC supply is distributed to multiple power rails, supplying \SI{12}{\volt} and \SI{24}{\volt} DC, as well as 230V AC through appropriate conversions.}
		\label{fig:power_supply}
\end{figure}

To power the components, a sophisticated power management system has been installed. An overview is given in \Cref{fig:power_supply}.
Power is provided primarily by a \SI{10}{\kWh} \SI{24}{\volt} battery in the trunk. This provides both \SI{24}{\volt} and stabilised \SI{12}{\volt} DC networks.
These networks power various built-in sensors and peripherals and up to 10 external devices through XT60 or XT90 ports.
An inverter provides \SI{230}{\volt} AC power for the compute unit, the network switch, the display matrix, and other external components connected via a \SI{230}{\volt} rail within the vehicle. 
During regular usage, the battery system demonstrated an average discharge of approximately \SI{1.3}{\kilo\watt}, translating to nearly eight hours of operation.
During a stress test with full compute load, and all components running, the system consumed \SI{2.3}{\kilo\watt} providing over four hours of runtime. To extend this runtime further, two \SI{600}{\watt} boosters draw power from the vehicle's \SI{12}{\volt} network if available. Thereby, the combustion engine as well as the hybrid battery can be utilized for additional operation time during mobile operations. For stationary operation, two external power connectors are integrated into the vehicle rear bumper. The first connector charges the component battery, providing up to \SI{3}{\kWh} during testing.
This ensures that, even under full demand, the system can operate indefinitely when connected to a power outlet.
The second connector directly powers the on-board \SI{230}{\volt} AC rail via the inverter pass-through to enable fast recharging even under high system load.
The system allows hot-plugging for seamless transition between stationary and mobile operation.
Moreover, all switches are routed through a programmable controller. This enables us to adapt to future upgrades by programming the switches to switch different component groups without the need for rewiring. Finally, the flexibility is further enhanced with a modular, IP66 certified roof wire feed-through. All power and data connections running to the components mounted on the roof rack are routed through this feed-through. Spare ports are available to mount further components in the future.

\section{DIGITAL TWIN AND SAMPLE DATA}
\label{sec:results}

To demonstrate the quality of the platform design, we provide a repository containing a digital twin of the vehicle. It provides accurate sensor positions and parameters as well as general vehicle data.
This digital twin is used in a CARLA simulation.
Moreover, we provide sample data of all of our sensors except radar. This enables others to evaluate the quality of our sensor suite. The data, together with the vehicle parameters, can be accessed here: \url{https://zenodo.org/records/11004676}.

\begin{figure}[t]
        \includegraphics[width=\columnwidth]{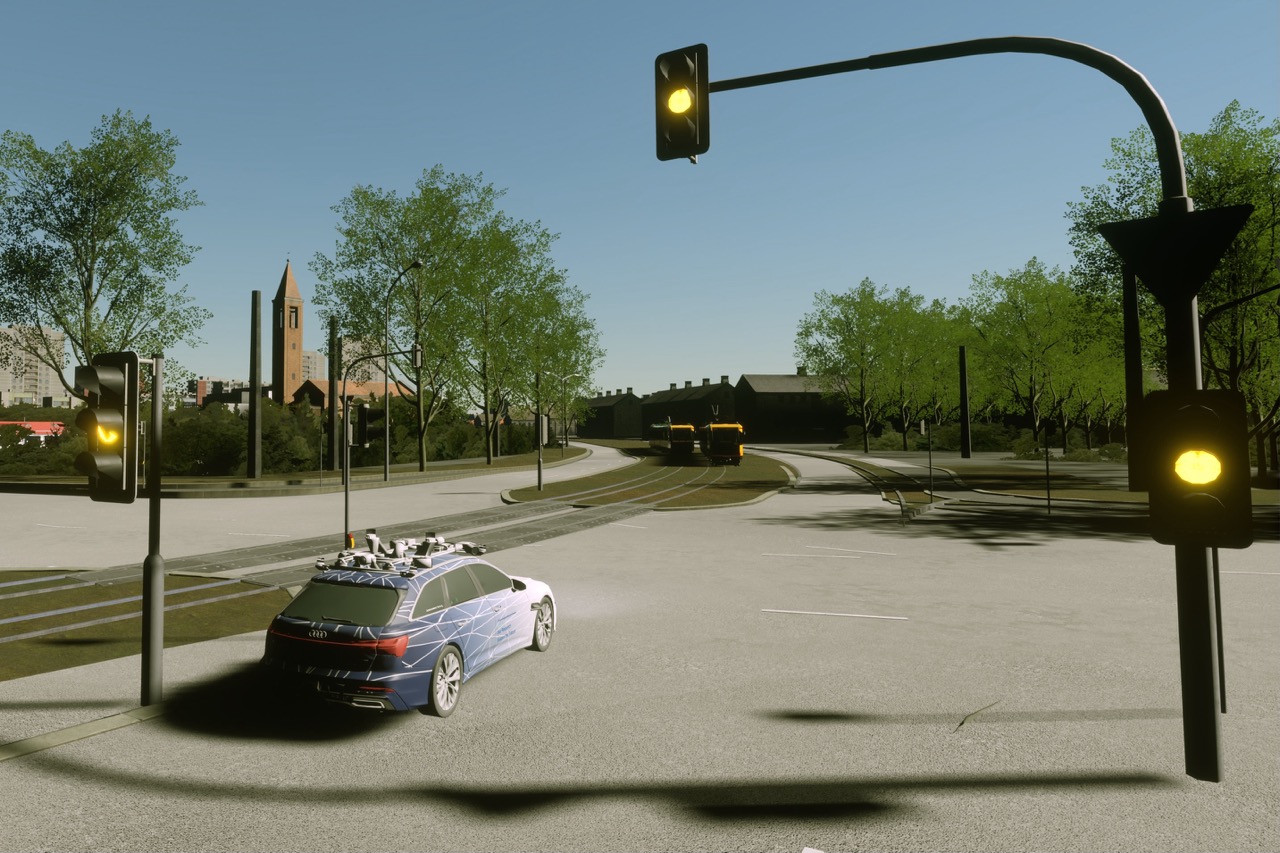}
    \caption{Digital twin of CoCar NextGen in a Carla simulation}
    \label{fig:cc-ng-sim}
\end{figure}

\section{CONCLUSIONS}
\label{sec:conclusion}

In this work we presented our modular, future-proof research vehicle for automated driving. Taking into account a multitude of use cases and scenarios, we built a extendable platform that covers a wide range of research needs. By integrating an extensive multi modal sensor suite with reference quality, together with a road-approved drive-by-wire system, a vital contribution was made to evaluate software for highly automated driving in real-world scenarios. Finally, we provided vehicle data together with sample data from the sensors for others to evaluate the quality of the setup. In future work we are targeting to provide a labelled dataset.

\section*{ACKNOWLEDGMENT}

The research leading to these results is funded by the German Federal Ministry for Economic Affairs and Climate Action within the ”Verbundprojekt: VVMethoden” (19A19002L) project and the ”Ministerium für Wirtschaft, Arbeit und Tourismus Baden-W\"urttemberg”.

\printbibliography

\end{document}